\documentclass{article}
\PassOptionsToPackage{numbers}{natbib}
\usepackage[preprint]{neurips_2026}
\usepackage[utf8]{inputenc}
\usepackage[T1]{fontenc}
\usepackage{hyperref}
\usepackage{url}
\usepackage{booktabs}
\usepackage{multirow}
\usepackage{graphicx}
\graphicspath{{comp/}}
\usepackage{amsfonts}
\usepackage{nicefrac}
\usepackage{microtype}
\usepackage{xcolor}
\usepackage{subcaption}
\usepackage{lipsum}
\usepackage{enumitem}
\usepackage{algorithm}
\usepackage{algorithmic}
\usepackage{chemformula}
\usepackage{pgfplots}
\pgfplotsset{compat=1.18}
\definecolor{plotblue}{HTML}{076678}
\definecolor{plotred}{RGB}{204,36,29}
\definecolor{gruvorange}{HTML}{af3a03}
\definecolor{gruvaqua}{HTML}{427b58}
\definecolor{gruvpurple}{HTML}{8f3f71}
\definecolor{gruvyellow}{HTML}{b57614}
\definecolor{gruvgreen}{HTML}{79740e}
\definecolor{gruvbrightorange}{HTML}{d65d0e}
\definecolor{gruvbrightaqua}{HTML}{689d6a}
\definecolor{gruvgray}{HTML}{665c54}

\hypersetup{
    colorlinks=true,
    linkcolor=plotblue,
    citecolor=plotblue,
    urlcolor=plotblue
}

\newtheorem{definition}{Definition}

\newcommand{\archname}[0]{\textsc{AMGenC}}
\newcommand{\std}[1]{{\scriptsize$\pm$#1}}
\newcommand{\first}[1]{\textcolor{gruvorange}{\textbf{#1}}}
\newcommand{\second}[1]{\textcolor{plotblue}{\underline{#1}}}

\usepackage{amsmath,amsfonts,bm}

\def\Secref#1{Section~\ref{#1}}

\def\eqref#1{equation~\ref{#1}}
\def\Eqref#1{Equation~\ref{#1}}

\def\Algref#1{Algorithm~\ref{#1}}

\def\1{\bm{1}}

\def\vzero{{\bm{0}}}
\def\vone{{\bm{1}}}

\def\vc{{\bm{c}}}

\def\ve{{\bm{e}}}

\def\vv{{\bm{v}}}

\def\vy{{\bm{y}}}

\def\mE{{\bm{E}}}

\def\mH{{\bm{H}}}
\def\mI{{\bm{I}}}

\def\mL{{\bm{L}}}

\def\mX{{\bm{X}}}

\DeclareMathAlphabet{\mathsfit}{\encodingdefault}{\sfdefault}{m}{sl}
\SetMathAlphabet{\mathsfit}{bold}{\encodingdefault}{\sfdefault}{bx}{n}

\def\gE{{\mathcal{E}}}

\def\gG{{\mathcal{G}}}

\def\gM{{\mathcal{M}}}
\def\gN{{\mathcal{N}}}

\def\gV{{\mathcal{V}}}

\def\sR{{\mathbb{R}}}

\def\sZ{{\mathbb{Z}}}

\newcommand{\E}{\mathbb{E}}
\newcommand{\Ls}{\mathcal{L}}

\newcommand{\softmax}{\mathrm{softmax}}

\DeclareMathOperator*{\argmax}{arg\,max}

\title{\archname: Generating Charge Balanced\\Amorphous Materials}

\author{%
Yan Lin\textsuperscript{\rm 1},~
Jilin Hu\textsuperscript{\rm 2}\thanks{Corresponding author.},~
N. M. Anoop Krishnan\textsuperscript{\rm 3,4},~
Morten M. Smedskjaer\textsuperscript{\rm 5}\\
\textsuperscript{\rm 1} Department of Computer Science, Aalborg University, Denmark\\
\textsuperscript{\rm 2} School of Data Science and Engineering, East China Normal University, China\\
\textsuperscript{\rm 3} Department of Civil Engineering, Indian Institute of Technology Delhi, India\\
\textsuperscript{\rm 4} Yardi School of Artificial Intelligence, Indian Institute of Technology Delhi, India\\
\textsuperscript{\rm 5} Department of Chemistry and Bioscience, Aalborg University, Denmark\\
\texttt{lyan@cs.aau.dk},\,
\texttt{jlhu@dase.ecnu.edu.cn},\,
\texttt{krishnan@iitd.ac.in},\,
\texttt{mos@bio.aau.dk}
}

\begin{document}

\maketitle

\begin{abstract}
Amorphous (disordered) materials require simulation cells with at least hundreds to thousands of atoms to accurately capture their diverse local atomic environments, compared to crystalline materials that can be described by unit cells containing a few to hundreds of atoms.
To advance the design of amorphous materials with desired properties and facilitate the exploration of their vast design space, generative inverse design has emerged as a promising approach. It aims to directly output materials with properties closely aligned with the desired ones using probabilistic generative models conditioned on desired properties, which can be more resource efficient than the traditional trial-and-error approach.
However, due to the inherent stochasticity of probabilistic generative models, when element assignments are unconstrained, a large portion of generated materials may be charge unbalanced, and existing methods either fix the composition in advance or discard charge unbalanced samples after generation.
In this work, we propose \archname, a new generative inverse design method for amorphous materials that can guarantee the generation of formally charge balanced samples, with minimal additional computational overhead and without sacrificing inverse design accuracy.
\archname{} achieves this through an element noise that gives the generation process a starting point centered around charge balance, and the combination of a per-step soft projection and a final discrete projection for steering the elements toward exact charge balance throughout the generation.
We perform extensive experiments on two amorphous materials datasets. Experimental results provide evidence that \archname{} achieves its design goal.
\end{abstract}

\section{Introduction}
Glasses and other amorphous materials are solids that lack a periodic atomic arrangement or long-range atomic order, yet exhibit complex short- and medium-range order.
They have shown great potential in domains including energy storage, thermal management, and advanced materials~\citep{liu2024amorphous}.
Formally, an amorphous material sample $\gM=(\mL, \mX, \mE)$ consists of a lattice $\mL$, atom positions $\mX$, and element assignments $\mE$ (Definition~\ref{def:amorphous-material}).
While the periodic structure of crystalline materials allows them to be fully described by a small cell containing only a few to hundreds of atoms, amorphous materials require larger cells with at least hundreds or thousands of atoms to accurately capture their diverse local atomic environments.

To advance the design of amorphous materials with desired properties, generative inverse design has emerged as a promising approach.
Compared to traditional materials design that relies on a resource-intensive trial-and-error process, it starts with the target properties, typically represented as a vector $\vy \in \sR^{n_y}$ of $n_y$ numerical properties, and directly outputs the sample $\gM$ with properties closely matched to the target $\vy$ using probabilistic generative models~\citep{DBLP:journals/corr/KingmaW13,goodfellow2014generative,DBLP:conf/nips/HoJA20,DBLP:conf/iclr/0011SKKEP21,DBLP:conf/iclr/LipmanCBNL23}.
Essentially, it can be formulated as a conditional probabilistic model $p_\theta (\gM|\vy)$, where $\theta$ represents the learnable parameters.
This approach has shown success in designing crystalline materials and molecules~\citep{gebauer2019symmetry,noh2019inverse,long2021constrained,court20203,DBLP:conf/iclr/XieFGBJ22,zeni2025generative,DBLP:conf/nips/WuGL0022,DBLP:conf/icml/HoogeboomSVW22,sriram2024flowllm,ye2024cdvae,luo2025crystalflow}.
Due to computational challenges in scaling generative models to hundreds or thousands of atoms, and the scarcity of diverse large-scale amorphous material datasets, generative inverse design remains relatively under-developed for amorphous materials~\citep{kwon2024spectroscopy,yang2025generative,finkler2025inverse}.
Nevertheless, the vast and largely unexplored design space of amorphous materials makes generative inverse design particularly promising, as it can efficiently navigate regions that are intractable for the traditional trial-and-error approach.

One limitation of existing generative inverse design methods is their inability to enforce charge balance constraints on the elements $\mE$ of generated samples without affecting the generative model's flexibility in adjusting the elements.
Due to the stochastic nature of probabilistic generative models, there is a certain probability of them generating charge unbalanced samples when $\mE$ is unconstrained.
Here, charge balance refers to formal charge balance, i.e., zero total charge under a supplied formal charge assignment $\vc$ (Definition~\ref{def:charge}).
For ionic oxide glasses such as those studied in this work, formal charge balance is equivalent to a sample containing exactly the oxygen needed to form neutral oxide units with its cations (Appendix~\ref{apx:stoichiometry}).
It is thus a necessary global condition on the composition of every sample, while local bonding, stability, and synthesizability require separate assessments.
A concrete use case is property-conditioned screening of glasses with variable composition, such as lithium-containing oxide glasses for solid electrolytes~\citep{ding2024amorphous,kalnaus2023solid}, where the target properties should steer both the composition and the structure.
Some existing methods~\citep{kwon2024spectroscopy,yang2025generative} fix the composition, i.e., $\mE$ is pre-defined and fixed during the generation, which limits the generation of materials with diverse compositions.
Other existing methods~\citep{DBLP:conf/iclr/XieFGBJ22,zeni2025generative,finkler2025inverse} follow the post-hoc filtering approach, i.e., discarding all charge unbalanced samples after the generation.
Post-hoc filtering is tolerable for crystalline materials, thanks to their small system scale, which usually results in less than 20\% of the generated samples being discarded~\citep{sriram2024flowllm,ye2024cdvae,luo2025crystalflow}.
However, it can be problematic for amorphous materials, since each sample has hundreds to thousands of atoms, often resulting in more than 90\% of the generated samples being discarded (\Secref{sec:comparison-methods}).
This defies the benefits of inverse design to some extent, since we would need to generate more samples to increase the chance of getting a charge balanced sample, effectively returning to the trial-and-error approach.

We identify that the main challenge in enforcing charge balance on $\mE$ comes down to the fact that it is a non-differentiable constraint, since computing $\mE$ involves $\argmax$ for probabilistic generative models.
While there are techniques for guiding the generation towards a non-differentiable preference~\citep{bansal2023universal,huang2024symbolic,yeh2024training}, they improve rather than guarantee the satisfaction of the preference.
Techniques for enforcing hard constraints either require the constraint to be differentiable~\citep{utkarsh2025physics} or restrict the action space of a sequential generator~\citep{hernandez2023crystal}, and are not directly applicable to discrete element assignments that evolve jointly with atom positions in a single flow.

To this end, we propose \textit{\underline{A}morphous \underline{M}aterial \underline{Gen}eration with \underline{C}harge Balanced Constraint} (\textbf{\archname{}}), a new method for generative inverse design of amorphous materials that can guarantee the generation of charge balanced samples.
Our contributions are as follows:
\begin{itemize}[leftmargin=*]
  \item We propose \archname{}, a flow-matching-based~\citep{DBLP:conf/iclr/LipmanCBNL23} generative inverse design method for amorphous materials that guarantees formally charge balanced generation with minimal computational overhead and without sacrificing inverse design accuracy.
  \item To achieve this, \archname{} introduces three components: a \textit{data-centered conditional element noise} that gives the generation a starting point centered around charge balance, a \textit{per-step soft linearized projection} that steers elements toward charge balance at each generation step by relaxing the non-differentiable constraint, and a \textit{final discrete projection} that resolves any residual charge imbalance through minimum-cost element swaps via dynamic programming. The first two components keep the charge residual small, so that the final discrete projection changes less than one element assignment per sample on average.
  \item We conduct extensive experiments on two amorphous materials datasets across multiple conditioning configurations. Results show that \archname{} guarantees charge balanced generation while matching or exceeding existing methods in inverse design accuracy, reducing the time to obtain the same amount of charge balanced samples by up to two orders of magnitude.
\end{itemize}

\section{Related Works}
\paragraph{Generative inverse design of materials.}
Traditional materials design~\citep{liu2017materials,cai2020machine} involves resource-intensive trial-and-error, where many samples are created in laboratories or simulation environments and have their properties tested.
Recent years have witnessed efforts on the generative inverse design approach aiming to be a more efficient alternative, where a conditional generative model directly generates materials with properties closely aligned with the target.
Early methods were based on two classic generative models, variational auto-encoder (VAE)~\citep{DBLP:journals/corr/KingmaW13} and generative adversarial network (GAN)~\citep{goodfellow2014generative}.
VAE-based methods~\citep{gebauer2019symmetry,noh2019inverse,court20203} have limited effectiveness in generating complex atomic systems~\citep{DBLP:conf/iclr/DaunhawerSCPV22}, while GAN-based methods~\citep{long2021constrained} are hampered by the unstable training of GANs~\citep{li2018limitations}.
More recent works based on diffusion models~\citep{DBLP:conf/nips/HoJA20,DBLP:conf/iclr/0011SKKEP21} or flow-matching~\citep{DBLP:conf/iclr/LipmanCBNL23} have shown promising results.
Most of them focus on relatively small atomic systems, i.e., crystalline materials~\citep{DBLP:conf/iclr/XieFGBJ22,zeni2025generative,sriram2024flowllm,ye2024cdvae,luo2025crystalflow} and molecules~\citep{DBLP:conf/nips/WuGL0022,DBLP:conf/icml/HoogeboomSVW22}, while there are only a few studies on amorphous materials~\citep{yang2025generative,finkler2025inverse}.
For crystalline materials, several methods first generate or select a charge balanced composition and then generate a structure conditioned on it~\citep{dong2024generative,yang2024genms,cheng2026enhancing}.
This is effective when the composition can be fixed before structure generation, and is complementary to our setting, where element assignments and atom positions evolve jointly in one property-conditioned generation process.
Beyond such composition-first pipelines, \textit{enforcing charge balance on generated samples} has not been extensively discussed in existing efforts for inverse design of materials.
We believe this is mainly due to the technical challenge of enforcing such non-differentiable constraints, and existing techniques for guiding or constraining generative models are ineffective or not directly applicable in this context.

\paragraph{Guiding or constraining generative models.}
Classifier guidance~\citep{dhariwal2021diffusion} and classifier-free guidance~\citep{ho2022classifier} are two representative techniques for guiding generative models with a preference for the generated data.
Classifier guidance requires the classifier to be differentiable. Alternatively, for non-differentiable classifiers or black-box preferences, gradient estimation~\citep{bansal2023universal} or stochastic search~\citep{huang2024symbolic,yeh2024training} techniques can be used.
Nonetheless, guidance techniques improve the satisfaction of a preference rather than guarantee an exact constraint on the final discrete output, such as the charge balance constraint in our context.
Crystal-GFN~\citep{hernandez2023crystal} imposes hard physical and geometrical constraints by restricting the action space of a sequential crystal generator, a generation process different from the joint flow of positions and elements considered here.
Physics-Constrained Flow Matching (PCFM)~\citep{utkarsh2025physics} introduces a solution for enforcing hard constraints on flow-matching. It can be applied to generation of continuous data, with a differentiable constraint expressed in an equality form.
In our context, however, $\mE$ is discrete, and the charge balance constraint is non-differentiable, making PCFM not directly applicable.

\section{Preliminaries}

\begin{definition}[Amorphous Material Sample]
\label{def:amorphous-material}
An amorphous material sample is represented as the positions and elements of atoms inside a unit cell, formally a tuple $\gM=(\mL, \mX, \mE)$ of three matrices.
$\mL \in \sR^{3\times 3}$ is the lattice of the unit cell, $\mX \in \sR^{n_a\times 3}$ is the positions of $n_a$ atoms, and $\mE \in \sR^{n_a\times d_E}$ is the one-hot embeddings of atomic elements, where $d_E$ is the number of element types under consideration.
The set of $n_y$ relevant properties is represented as $\vy\in \sR^{n_y}$, where each value denotes the magnitude of that property.
\end{definition}

\begin{definition}[Charge]
\label{def:charge}
Each element type $j \in \{1, \ldots, d_E\}$ has an associated formal charge value $c_j \in \sZ$, collected into a vector $\vc \in \sZ^{d_E}$.
The total charge of a sample $\gM$ is defined as $Q(\mE) = \vone^\top \mE\, \vc$,
where $\vone \in \sR^{n_a}$ is the all-ones vector.
A sample is said to be charge balanced if $Q(\mE) = 0$.
This formal charge balance does not by itself imply electronic neutrality or local chemical validity.
\end{definition}

\begin{definition}[Ghost Atoms]
Ghost atoms are introduced to allow a generative model to control the density of samples without modifying $\mL$ or the number of atoms.
Formally, they are a special atom type incorporated into each sample.
In the datasets, ghost atoms are randomly positioned within the cell so that the total number of atoms is $n_a=\lfloor \rho\cdot\text{Vol}(\mL) \rfloor$, where $\rho$ is the maximum density.
Ghost atoms are treated like normal atoms by the model but are assigned a special chemical element class with formal charge set to 0.
The model adjusts sample density by changing the fraction of ghost atoms, which are removed after generation.
\end{definition}

\noindent \textbf{Problem definition.}
\textit{Inverse design of amorphous materials with charge balance constraint.}
Given a set of desired properties $\vy$, the goal is to learn a conditional probabilistic model $p_\theta(\gM|\vy)$ that generates amorphous material samples $\gM$ with properties closely aligned with $\vy$, subject to the constraint that all generated samples are charge balanced, i.e., $Q(\mE) = 0$.

\section{\archname{} Architecture}
\label{sec:method}

\archname{} is built on conditional flow-matching~\citep{DBLP:conf/iclr/LipmanCBNL23} (\Secref{sec:fm}).
To enforce charge balance, it introduces three components acting at different stages of the generation pipeline:
a \textit{data-centered conditional element noise} (\Secref{sec:ot-noise}),
a \textit{per-step soft linearized projection} (\Secref{sec:pcfm}),
and a \textit{final discrete projection} (\Secref{sec:dp}).

\begin{figure}[t]
  \centering
  \begin{subfigure}[b]{0.24\textwidth}
    \centering
    \includegraphics[width=\textwidth]{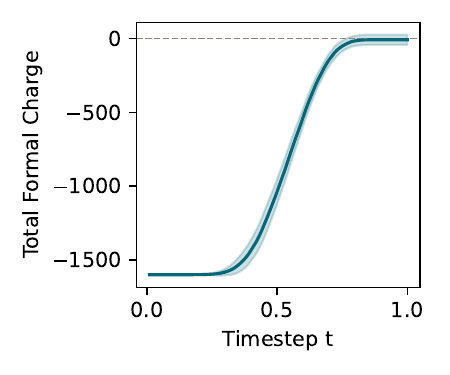}
    \caption{Standard flow matching}
    \label{fig:convergence-base}
  \end{subfigure}
  \hfill
  \begin{subfigure}[b]{0.24\textwidth}
    \centering
    \includegraphics[width=\textwidth]{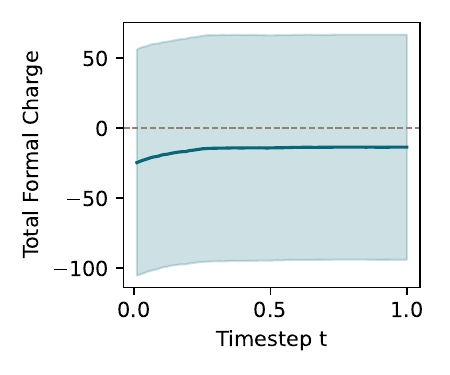}
    \caption{Add element noise}
    \label{fig:convergence-noise}
  \end{subfigure}
  \hfill
  \begin{subfigure}[b]{0.24\textwidth}
    \centering
    \includegraphics[width=\textwidth]{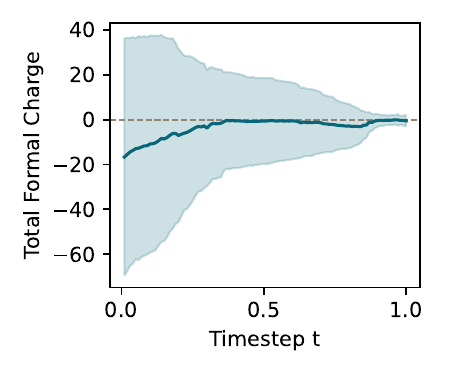}
    \caption{Add soft projection}
    \label{fig:convergence-proj}
  \end{subfigure}
  \hfill
  \begin{subfigure}[b]{0.24\textwidth}
    \centering
    \includegraphics[width=\textwidth]{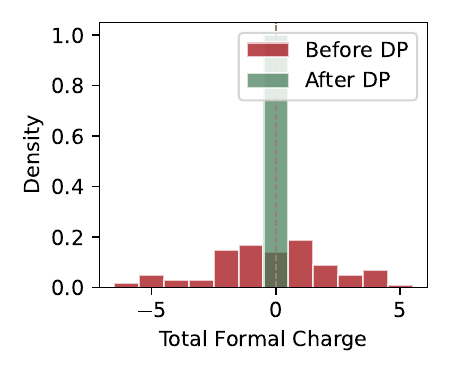}
    \caption{Add final projection}
    \label{fig:charge-dist}
  \end{subfigure}
  \caption{Illustration of how each component in \archname{} progressively reduces the charge residual during generation. (a)--(c): total charge $Q(\hat{\mE}_1)$ over generation steps as components are added cumulatively. (d): distribution of total charge before and after the final discrete projection (DP).}
  \label{fig:ablation}
\end{figure}

\subsection{Conditional Flow Matching}
\label{sec:fm}

\archname{} uses flow-matching with linear interpolation paths.
We define time $t \in [0, 1]$, where $t=0$ corresponds to noise and $t=1$ to clean data.
During the flow, both positions $\mX$ and element embeddings $\mE$ evolve in continuous space, while the lattice $\mL$ is fixed.
The flow paths are:
\begin{align}
\mX_t &= \mX_0 + t \cdot \vv_\mX \label{eq:flow-pos}\\
\mE_t &= \mE_0 + t \cdot \vv_\mE \label{eq:flow-el}
\end{align}
where $\mX_0$ is sampled uniformly in the unit cell, $\mE_0$ is a source noise (discussed in \Secref{sec:ot-noise}),
$\vv_\mX$ is the displacement from $\mX_0$ to $\mX_1$ computed under periodic boundary conditions, $\vv_\mE = \mE_1 - \mE_0$,
and $\mX_1$ and $\mE_1$ are from the training data.

The velocity predictor $\vv_\theta(\gM_t, \vy, t) = (\vv_\mX, \vv_\mE)$ is parameterized by an E(n)-equivariant graph neural network (EGNN)~\citep{DBLP:conf/icml/SatorrasHW21}.
Input node features are the concatenation of the scalar time $t$, the current element embedding, and a linear projection of the target property vector $\vy$.
Edges connect atom pairs within a cutoff radius $r_\text{cut}$, and the network outputs a position velocity $\vv_\mX \in \sR^{n_a \times 3}$ and an element velocity $\vv_\mE \in \sR^{n_a \times d_E}$.
Details of the EGNN architecture are provided in Appendix~\ref{apx:egnn}.

The training objective minimizes the mean squared error between predicted and target velocities:
\begin{equation}
\Ls = \E_{t,\, \gM} \left[ \| \vv_{\mX,\theta} - \vv_\mX \|^2 + \| \vv_{\mE,\theta} - \vv_\mE \|^2 \right]
\label{eq:loss}
\end{equation}
where $t \sim \mathrm{Uniform}(0,1)$, $\gM \sim p_\text{data}$.

During inference, generation proceeds by Euler integration from $t=0$ to $t=1$ with $T$ steps of size $\Delta t = 1/T$.
At each step, a clean extrapolation predicts the final element state:
\begin{equation}
\hat{\mE}_1 = \mE_t + (1 - t) \cdot \vv_{\mE,\theta}(\gM_t, \vy, t)
\label{eq:clean-extrap}
\end{equation}
where $\gM_t = (\mL, \mX_t, \mE_t)$ is the intermediate state.
The discrete element assignments are obtained from $\hat{\mE}_1$ via $\argmax$ at $t=1$.

In standard flow-matching, no mechanism exists to enforce constraints on the discrete element assignments.
The following three subsections address this.

\subsection{Data-Centered Conditional Element Noise}
\label{sec:ot-noise}

In standard flow-matching, element noise is sampled from an isotropic Gaussian $\mE_0 \sim \gN(\vzero, \mI)$.
Through $\softmax$, this produces a near-uniform distribution over elements.
However, in many material systems, the element distribution is highly imbalanced.
For example, in oxide glasses, oxygens typically constitute over 60\% of atoms, which becomes the fraction needed to offset the positive charges of other elements to achieve $Q = 0$.
The uniform noise distribution reduces oxygen's prevalence from ${\sim}60\%$ to ${\sim}1/d_E$, creating a large positive charge imbalance at $t=0$ that the model must spend early trajectory capacity correcting (Figure~\ref{fig:convergence-base}).

We address this with a \textit{data-centered conditional element noise}.
During training, the source noise is centered at the clean element matrix:
\begin{equation}
\mE_0 = \mE_1 + \sigma \bm{\epsilon},\quad \bm{\epsilon} \sim \gN(\vzero, \mI)
\label{eq:ot-noise}
\end{equation}
where $\sigma > 0$ is a noise scale hyperparameter.
This is a conditional coupling where each noise sample is paired with the data point it is centered on, which reduces path crossings similar to optimal transport coupling in flow-matching~\citep{tong2024improving}, although no batch transport plan or transport cost is computed.

Over the training set, this induces a Gaussian mixture as the per-atom marginal noise distribution:
\begin{equation}
p(\mE_{0,i}) = \sum_{k=1}^{d_E} f_k \cdot \gN(\mE_{0,i};\, \ve_k,\, \sigma^2 \mI)
\label{eq:marginal}
\end{equation}
where $f_k$ is the fraction of atoms assigned to element $k$ in the training set.
At inference, since the true elements are unknown, we sample each row of $\mE_0$ independently from this marginal by drawing $k_i \sim \mathrm{Cat}(f_1, \ldots, f_{d_E})$ and $\mE_{0,i} \sim \gN(\ve_{k_i}, \sigma^2 \mI)$, matching the training distribution and keeping the source close to charge balance.

The above element noise provides a better starting point (Figure~\ref{fig:convergence-noise}) but does not guarantee charge balance on its own.
The following two components actively steer and correct the elements during and after generation.

\subsection{Per-Step Soft Linearized Projection}
\label{sec:pcfm}

During inference, we apply a projection at each step to steer the predicted $\hat{\mE}_1$ (\Eqref{eq:clean-extrap}) toward charge balance.
Inspired by the Gauss-Newton projection in PCFM~\citep{utkarsh2025physics}, we move the predicted clean elements toward the constraint $Q = 0$ at each inference step with a minimum-norm correction for the linearized constraint.
For a scalar constraint like ours, this correction has a simple closed form that finds the smallest change to the element logits needed to satisfy the linearized constraint, making it both efficient and minimally disruptive to the model's predictions (\Secref{sec:model-analysis}).
It does, however, require the constraint to be differentiable.
In our setting, charge balance involves discrete element assignments: $Q = \vone^\top \mathrm{onehot}(\argmax(\hat{\mE}_1))\, \vc$, which makes $Q$ non-differentiable.

We make two design choices to handle this non-differentiability.
First, we relax the $\argmax$ with $\softmax(\hat{\mE}_1 / \tau)$ at temperature $\tau$, yielding a differentiable soft charge whose gradient can be computed via backpropagation:
\begin{equation}
Q_\text{soft} = \vone^\top \softmax(\hat{\mE}_1 / \tau)\, \vc
\label{eq:q-soft}
\end{equation}

Second, we decouple the charge value from its gradient, because no single $\tau$ serves both well:
at high $\tau$, $Q_\text{soft}$ drifts from the true discrete charge $Q$;
at low $\tau$, $Q_\text{soft}$ is accurate but its gradient vanishes due to softmax saturation, causing the projection to be unstable.
We therefore use the exact discrete charge $Q$ as the residual, and compute the gradient $\nabla_{\hat{\mE}_1} Q_\text{soft}$ through $\softmax(\hat{\mE}_1 / \tau)$ at a moderate $\tau$ as the Jacobian of the relaxed constraint.

The soft linearized projection then corrects $\hat{\mE}_1$:
\begin{equation}
\hat{\mE}_1^\text{proj} = \hat{\mE}_1 - \frac{Q}{\|\nabla_{\hat{\mE}_1} Q_\text{soft}\|^2}\,\nabla_{\hat{\mE}_1} Q_\text{soft}
\label{eq:gn-projection}
\end{equation}
Since the residual comes from the discrete charge $Q$ and the Jacobian from its relaxation $Q_\text{soft}$, this is not an exact Gauss-Newton step, but a minimum-norm correction for the linearized relaxed constraint.
The gradient $\nabla_{\hat{\mE}_1} Q_\text{soft}$ is largest at atoms where the model is least confident about the element assignment, so the correction preferentially adjusts these atoms while leaving confident ones unchanged.

After projection, we re-interpolate along the flow path in \Eqref{eq:flow-el} with the corrected target:
\begin{equation}
\bar{\mE}_{t'} = (1 - t')\,\mE_0 + t'\,\hat{\mE}_1^\text{proj}
\label{eq:ot-interpolant}
\end{equation}
which gives the corrected starting state $\bar{\mE}_{t'}$ for the next step at $t' = t + \Delta t$.
This per-step correction progressively tightens the charge convergence over the generation trajectory (Figure~\ref{fig:convergence-proj}).
The full inference procedure is given in \Algref{alg:inference} in Appendix~\ref{apx:inference-algorithm}.

\subsection{Final Discrete Projection}
\label{sec:dp}

After the flow reaches $t=1$, the element assignments from $\argmax$ may still have $Q \neq 0$, since both the softmax relaxation and the use of estimated $\hat{\mE}_1$ at each step can introduce error.
To guarantee exact charge balance, we apply a discrete projection as the last step of generation, finding the minimum-cost set of element swaps to achieve $Q = 0$ (Figure~\ref{fig:charge-dist}).

Given the element logits $\hat{\mE}_1$ at $t=1$ and current element assignments $e_i = \argmax_j \hat{\mE}_{1,ij}$ of the $i$-th atom, we solve:
\begin{equation}
\min_{\{e_i'\}_{i=1}^{n_a}} \sum_{i=1}^{n_a} \left(\hat{\mE}_{1,i,e_i} - \hat{\mE}_{1,i,e_i'}\right) \quad \text{s.t.} \quad \sum_{i=1}^{n_a} c_{e_i'} = 0
\label{eq:dp-objective}
\end{equation}
where $c_j$ is the formal charge of element $j$.
The cost $\hat{\mE}_{1,i,e_i} - \hat{\mE}_{1,i,e_i'}$ is the logit gap sacrificed by swapping atom $i$ from $e_i$ to $e_i'$, ensuring that atoms where the model is most confident are swapped last.

Since formal charges are integers, the charge residual $Q$ lives in a bounded integer range, making it suitable as a state variable for dynamic programming.
Our implementation of the algorithm (detailed in \Algref{alg:dp} in Appendix~\ref{apx:dp-algorithm}) processes atoms one by one, considering all possible element assignments at each atom while tracking the accumulated charge delta and total swap cost, to find the assignment that reaches $Q = 0$ with minimum cost.

The complexity is $O(n_a \cdot |Q_\text{max}| \cdot d_E)$, where $|Q_\text{max}|$ is the range of achievable charge values.
In practice, after the soft linearized projection has already reduced the charge residual to a small range, the algorithm takes approximately 1\% of total inference time while guaranteeing $Q = 0$ for every generated sample (\Secref{sec:model-analysis}).

\section{Experiments}
We evaluate the inverse design performance of \archname{} on two amorphous material datasets, against several existing generative inverse design methods for materials.

\subsection{Datasets}
The two datasets~\citep{finkler2025inverse} are obtained using classical molecular dynamics simulation workflows based on LAMMPS~\citep{lammps} and ASE~\citep{larsen2017atomic}.
First, the \textbf{amorphous silica (a-\ch{SiO2}) dataset}, which contains silica (\ch{SiO2}) samples that vary in size (between 80 and 250 atoms) and whose properties depend on structures and densities, since the composition is fixed.
Second, the \textbf{multi-element glass (MEG) dataset}, which consists of 9,027 glass samples, each containing approximately 800 atoms across 11 different elements. Initial structures were generated from varying compositions of the glass network formers \ch{SiO2} and \ch{P2O5}, and the network modifiers \ch{Al2O3}, \ch{Li2O}, \ch{BeO}, \ch{K2O}, \ch{CaO}, \ch{TiO2}, \ch{BaO} and \ch{ZnO}.
More details about the preparation of the datasets are provided in Appendix~\ref{apx:datasets}.

\subsection{Comparison Methods}
We compare \archname{} with several generative methods for inverse design of materials.
\textbf{CDVAE}~\citep{DBLP:conf/iclr/XieFGBJ22} is a diffusion-based VAE that generates stable crystalline materials by denoising atomic coordinates and types through Langevin dynamics.
\textbf{CrystalFlow}~\citep{luo2025crystalflow} is a flow-based model that generates crystalline materials by learning continuous normalizing flows with conditional flow matching.
\textbf{MatterGen}~\citep{zeni2025generative} is a diffusion model that generates crystalline materials across the periodic table and supports fine-tuning for property-guided generation.
\textbf{Graphite}~\citep{kwon2024spectroscopy} is a diffusion model that generates amorphous materials using XANES spectroscopy as conditioning for inverse structure prediction.
\textbf{AMDEN}~\citep{finkler2025inverse} is a diffusion model that generates amorphous materials with optimization for low-energy structures by incorporating relaxation into the diffusion generation process.
All baselines are retrained on the same datasets and conditioned on the same target properties as \archname{}, with the adaptation of each baseline detailed in Appendix~\ref{apx:baseline-adaptation}.

\subsection{Experimental Settings}
\label{sec:experimental-settings}
We evaluate the models' inverse design performance by performing generation conditioned on specific target properties, and compare the actual properties of generated samples with the target.
We also evaluate the charge balance of generated samples under each configuration, as well as the computational efficiency implication.
Models are trained with the Adam optimizer (learning rate $10^{-3}$) for 200 epochs with batch size 8.
All methods use 100 sampling steps during inference for a fair comparison.
Each experiment is repeated 5 times, and the mean and standard deviation of each metric are reported.
Details of the computational environment are provided in Appendix~\ref{apx:environment}.

\paragraph{Settings for the a-\ch{SiO2} dataset.}
For the a-\ch{SiO2} dataset, we are interested in the following two properties of samples.
Shear modulus $G$ characterizes a material's resistance to elastic deformation under shear stress, representing mechanical stiffness. Ring size distribution (RSD) quantifies the medium-range order in amorphous silica by measuring the average number of \ch{Si} atoms in rings formed by the Si-O network.
We perform generation conditioned on either one of them, denoted as a-\ch{SiO2}~$\mid$~$G$ and a-\ch{SiO2}~$\mid$~RSD respectively, with target shear modulus linearly interpolated between 10 and 50 [GPa] and target RSD linearly interpolated between 4 and 6 atoms, both across 2,000 samples with cubic cells with edge length of 18\,\AA, and $\rho = 0.11~\text{atoms/\AA}^3$.
The formal charge values are $c_{\mathrm{Si}} = +4$ and $c_{\mathrm{O}} = -2$, with ghost atoms assigned $c = 0$, so $Q = 0$ is equivalent to the exact \ch{SiO2} stoichiometry $N_\mathrm{O} = 2N_\mathrm{Si}$ after removing ghost atoms.
Although a-\ch{SiO2} has a fixed composition, the two target properties are closely related to sample density, which is controlled via ghost atoms. The element configuration thus remains unfixed, and charge balance enforcement is still necessary.

\paragraph{Settings for the MEG dataset.}
For the MEG dataset, we are interested in the following three properties of the samples.
In addition to shear modulus $G$, Young's modulus $E$ can be used to characterize elastic properties. The lithium molar concentration $C_\text{Li}$ quantifies the composition ratio, which is important for applications related to, e.g., glassy solid electrolytes for batteries~\citep{ding2024amorphous,kalnaus2023solid}.
We perform generation conditioned on the combination of Young's modulus and lithium molar concentration, or on the shear modulus, denoted as MEG~$\mid$~$E$+$C_\text{Li}$ and MEG~$\mid$~$G$ respectively.
The target Young's modulus is linearly interpolated between 20 and 160 GPa, the target lithium molar concentration is fixed at 15\%, and the target shear modulus is linearly interpolated between 10 and 70 GPa, all across 2,000 samples with cubic cells with edge length of 23\,\AA, and $\rho = 0.11~\text{atoms/\AA}^3$.
The formal charge values for the 11 element types are derived from the BMP potential~\citep{bertani2021improved}. The full assignment is provided in Appendix~\ref{apx:charge-assignment}.
Under these formal charge values, $Q = 0$ is equivalent to a sample containing exactly the oxygen needed to form neutral oxide units with its cations, while the cation ratios remain free for the property-conditioned generation to steer, e.g., toward the target $C_\text{Li}$ (Appendix~\ref{apx:stoichiometry}).

For both datasets, all training samples are charge balanced (Appendix~\ref{apx:dataset-charge}), and the target ranges extend beyond the training data distributions, requiring the model to extrapolate (Appendix~\ref{apx:property-distributions}).

\paragraph{Evaluation metrics.}
We evaluate the inverse design performance by comparing target properties with properties of generated samples using three regression metrics: Mean Absolute Error (MAE), Root Mean Square Error (RMSE), and Mean Absolute Percentage Error (MAPE). Implementation details are provided in Appendix~\ref{apx:inverse-design-metrics}.
Detailed procedures for calculating the properties from generated samples are provided in Appendix~\ref{apx:property-calculation}.
For charge balance, we report three metrics over generated samples: the probability $P(Q{=}0)$ of charge balance, measured as the percentage of samples satisfying $Q(\mE) = 0$, and the mean absolute charge $|\bar{Q}|$ and standard deviation $\sigma_Q$ of the charge value $Q(\mE)$ (definitions in Appendix~\ref{apx:charge-metrics}).
For computational efficiency, we report the wall-clock time $T_{1\text{k}}$ to generate 1,000 samples, and the time $T_{1\text{k}}^{Q=0}$ to generate 1,000 charge-balanced samples, which accounts for the need to discard and regenerate samples that violate charge balance.

\subsection{Comparison of Methods}
\label{sec:comparison-methods}
\paragraph{Inverse design metrics comparison.}
Table~\ref{tab:inverse-design} reports the inverse design performance across all conditioning configurations. \archname{} achieves the best or second best result on every metric in all four configurations, ranking first on all metrics on the a-\ch{SiO2} dataset and trading the top position with AMDEN on the MEG dataset.
Since none of the baselines enforce charge balance, the comparable inverse design performance of \archname{} suggests that its components for enforcing charge balance do not sacrifice inverse design accuracy.
We further compare several structural features of the generated samples with those of the training samples in Appendix~\ref{apx:structural-features}.

\begin{table}[t]
  \caption{Inverse design performance comparison across conditioning configurations. Lower is better for all metrics. \first{Bold} and \second{underline} indicate the best and second best results, respectively.}
  \label{tab:inverse-design}
  \centering
  \setlength{\tabcolsep}{4.2pt}
  \resizebox{\textwidth}{!}{
\begin{tabular}{ll cccccc}
  \toprule
  Config & Metric & CDVAE & CrystalFlow & MatterGen & Graphite & AMDEN & \archname{} \\
  \midrule
  \multirow{3}{*}{a-\ch{SiO2} $\mid$ $G$}
    & MAE $\downarrow$  & 4.25\std{0.14} & 3.96\std{0.11} & 3.89\std{0.12} & 3.58\std{0.08} & \second{3.43\std{0.11}} & \first{3.32\std{0.10}} \\
    & RMSE $\downarrow$ & 5.18\std{0.16} & 4.81\std{0.15} & 4.72\std{0.14} & 4.35\std{0.10} & \second{4.21\std{0.12}} & \first{4.19\std{0.13}} \\
    & MAPE $\downarrow$ & 15.8\std{0.7}\% & 15.1\std{0.6}\% & 14.9\std{0.5}\% & \second{14.3\std{0.3}\%} & 15.2\std{0.6}\% & \first{14.1\std{0.4}\%} \\
  \cmidrule(lr){1-8}
  \multirow{3}{*}{a-\ch{SiO2} $\mid$ RSD}
    & MAE $\downarrow$  & 0.22\std{0.01} & 0.20\std{0.02} & 0.19\std{0.02} & \second{0.17\std{0.01}} & 0.17\std{0.01} & \first{0.16\std{0.01}} \\
    & RMSE $\downarrow$ & 0.27\std{0.02} & 0.25\std{0.01} & 0.24\std{0.02} & \second{0.21\std{0.01}} & 0.21\std{0.01} & \first{0.20\std{0.01}} \\
    & MAPE $\downarrow$ & 4.5\std{0.3}\% & 4.0\std{0.2}\% & 3.9\std{0.3}\% & 3.6\std{0.1}\% & \second{3.5\std{0.2}\%} & \first{3.4\std{0.2}\%} \\
  \cmidrule(lr){1-8}
  \multirow{6}{*}{MEG $\mid$ $E$+$C_\text{Li}$}
    & $E$ MAE $\downarrow$           & 13.02\std{0.38} & 11.65\std{0.29} & 11.32\std{0.33} & 10.15\std{0.21} & \first{9.85\std{0.27}} & \second{9.92\std{0.26}} \\
    & $E$ RMSE $\downarrow$          & 15.88\std{0.45} & 14.23\std{0.32} & 13.82\std{0.40} & 12.38\std{0.26} & \second{12.02\std{0.34}} & \first{11.82\std{0.29}} \\
    & $E$ MAPE $\downarrow$          & 26.3\std{0.9}\% & 23.6\std{0.6}\% & 22.9\std{0.8}\% & 20.5\std{0.5}\% & \first{19.9\std{0.6}\%} & \second{20.1\std{0.5}\%} \\
    & $C_\text{Li}$ MAE $\downarrow$  & 2.02\std{0.05} & 1.80\std{0.04} & 1.75\std{0.06} & 1.56\std{0.03} & \second{1.52\std{0.04}} & \first{1.49\std{0.04}} \\
    & $C_\text{Li}$ RMSE $\downarrow$ & 2.45\std{0.06} & 2.19\std{0.04} & 2.12\std{0.07} & 1.90\std{0.04} & \second{1.85\std{0.05}} & \first{1.82\std{0.04}} \\
    & $C_\text{Li}$ MAPE $\downarrow$ & 13.3\std{0.4}\% & 11.9\std{0.3}\% & 11.6\std{0.5}\% & 10.4\std{0.2}\% & \second{10.1\std{0.3}\%} & \first{9.8\std{0.3}\%} \\
  \cmidrule(lr){1-8}
  \multirow{3}{*}{MEG $\mid$ $G$}
    & MAE $\downarrow$  & 7.08\std{0.21} & 6.34\std{0.15} & 6.15\std{0.18} & 5.52\std{0.12} & \second{5.35\std{0.15}} & \first{5.32\std{0.14}} \\
    & RMSE $\downarrow$ & 8.75\std{0.25} & 7.85\std{0.19} & 7.62\std{0.22} & 6.82\std{0.15} & \second{6.62\std{0.19}} & \first{6.48\std{0.17}} \\
    & MAPE $\downarrow$ & 17.8\std{0.7}\% & 16.0\std{0.4}\% & 15.5\std{0.5}\% & 13.9\std{0.3}\% & \second{13.5\std{0.5}\%} & \first{13.3\std{0.4}\%} \\
  \bottomrule
\end{tabular}
}
\end{table}

\paragraph{Charge balance comparison.}
Table~\ref{tab:charge-balance} reports the charge balance metrics. \archname{} guarantees the generation of charge balanced samples across all configurations.
All baseline methods, where element assignments are unconstrained, produce charge balanced samples at low rates that decrease as the number of atoms and element types grows. On the a-\ch{SiO2} dataset with two element types and 80 to 250 atoms, the best baseline reaches at most 6.8\% $P(Q{=}0)$. On the MEG dataset with 11 element types and approximately 800 atoms, this drops to at most 2.0\%.
This trend suggests that for larger amorphous material systems, post-hoc filtering alone would discard more generated samples.
As none of the baselines enforce charge balance, this comparison quantifies how the cost of leaving the constraint unenforced grows with system scale.

\begin{table}[t]
  \caption{Charge balance comparison across conditioning configurations. Higher $P(Q{=}0)$ and lower $|\bar{Q}|$, $\sigma_Q$ are better. \first{Bold} and \second{underline} indicate the best and second best results, respectively.}
  \label{tab:charge-balance}
  \centering
  \setlength{\tabcolsep}{4.5pt}
  \resizebox{\textwidth}{!}{
\begin{tabular}{ll cccccc}
  \toprule
  Config & Metric & CDVAE & CrystalFlow & MatterGen & Graphite & AMDEN & \archname{} \\
  \midrule
  \multirow{3}{*}{a-\ch{SiO2} $\mid$ $G$}
    & $P(Q{=}0)$ $\uparrow$  & 3.6\std{0.5}\% & 3.1\std{0.4}\% & 4.2\std{0.6}\% & 5.2\std{0.8}\% & \second{5.5\std{0.8}\%} & \first{100.0\std{0.0}\%} \\
    & $|\bar{Q}|$ $\downarrow$  & 2.93\std{0.41} & 3.28\std{0.48} & 2.26\std{0.38} & \second{0.98\std{0.17}} & 1.15\std{0.22} & \first{0.00\std{0.00}} \\
    & $\sigma_Q$ $\downarrow$ & 10.47\std{1.23} & 11.35\std{1.42} & 9.03\std{0.95} & 7.52\std{0.84} & \second{7.31\std{0.76}} & \first{0.00\std{0.00}} \\
  \cmidrule(lr){1-8}
  \multirow{3}{*}{a-\ch{SiO2} $\mid$ RSD}
    & $P(Q{=}0)$ $\uparrow$  & 4.1\std{0.5}\% & 3.5\std{0.6}\% & 4.8\std{0.6}\% & \second{6.8\std{0.9}\%} & 6.4\std{0.7}\% & \first{100.0\std{0.0}\%} \\
    & $|\bar{Q}|$ $\downarrow$  & 6.24\std{0.73} & 6.92\std{0.78} & 5.38\std{0.64} & 3.92\std{0.48} & \second{3.73\std{0.45}} & \first{0.00\std{0.00}} \\
    & $\sigma_Q$ $\downarrow$ & 8.15\std{0.92} & 8.83\std{1.08} & 6.92\std{0.81} & 4.93\std{0.58} & \second{4.67\std{0.53}} & \first{0.00\std{0.00}} \\
  \cmidrule(lr){1-8}
  \multirow{3}{*}{MEG $\mid$ $E$+$C_\text{Li}$}
    & $P(Q{=}0)$ $\uparrow$  & 1.0\std{0.2}\% & 0.8\std{0.1}\% & 1.1\std{0.3}\% & 1.4\std{0.3}\% & \second{1.6\std{0.4}\%} & \first{100.0\std{0.0}\%} \\
    & $|\bar{Q}|$ $\downarrow$  & 26.58\std{3.84} & 38.15\std{5.45} & 24.73\std{3.12} & 21.06\std{2.85} & \second{18.84\std{2.51}} & \first{0.00\std{0.00}} \\
    & $\sigma_Q$ $\downarrow$ & 78.23\std{6.53} & 82.47\std{5.81} & 66.48\std{5.82} & \second{49.18\std{4.15}} & 50.95\std{4.38} & \first{0.00\std{0.00}} \\
  \cmidrule(lr){1-8}
  \multirow{3}{*}{MEG $\mid$ $G$}
    & $P(Q{=}0)$ $\uparrow$  & 1.3\std{0.3}\% & 1.0\std{0.2}\% & 1.4\std{0.3}\% & 1.8\std{0.4}\% & \second{2.0\std{0.4}\%} & \first{100.0\std{0.0}\%} \\
    & $|\bar{Q}|$ $\downarrow$  & 16.47\std{2.83} & 17.93\std{2.42} & 16.82\std{2.25} & \second{11.37\std{1.68}} & 11.91\std{1.76} & \first{0.00\std{0.00}} \\
    & $\sigma_Q$ $\downarrow$ & 67.14\std{5.72} & 72.38\std{5.13} & 57.86\std{5.04} & 46.53\std{3.91} & \second{45.22\std{3.85}} & \first{0.00\std{0.00}} \\
  \bottomrule
\end{tabular}
}
\end{table}

\paragraph{Computation efficiency comparison.}
Table~\ref{tab:efficiency} reports the computation efficiency.
While the generation time $T_{1\text{k}}$ is comparable across methods, the time to generate 1,000 charge balanced samples ($T_{1\text{k}}^{Q=0}$) reveals a large gap.
Baselines that rely on post-hoc filtering must generate and discard far more samples due to their low $P(Q{=}0)$. On the MEG dataset, where baselines discard over 98\% of generated samples, the best baseline requires approximately 39 to 49 hours, while \archname{} needs approximately 47 minutes.
This gap would widen for larger systems where $P(Q{=}0)$ is expected to be lower.

\begin{table}[t]
  \caption{Computation efficiency comparison across conditioning configurations. Lower is better for all metrics. \first{Bold} and \second{underline} indicate the best and second best results, respectively.}
  \label{tab:efficiency}
  \centering
  \setlength{\tabcolsep}{3.0pt}
  \resizebox{\textwidth}{!}{
\begin{tabular}{ll cccccc}
  \toprule
  Config & Metric & CDVAE & CrystalFlow & MatterGen & Graphite & AMDEN & \archname{} \\
  \midrule
  \multirow{2}{*}{a-\ch{SiO2} $\mid$ $G$}
    & $T_{1\text{k}}$ $\downarrow$       & 26.37\std{0.45}m & 28.42\std{0.47}m & 31.89\std{0.58}m & 23.74\std{0.31}m & \first{22.93\std{0.38}m} & \second{23.68\std{0.53}m} \\
    & $T_{1\text{k}}^{Q=0}$ $\downarrow$ & 12.21\std{1.71}h & 15.28\std{2.63}h & 12.65\std{1.82}h & 7.61\std{1.17}h & \second{6.95\std{1.02}h} & \first{23.68\std{0.53}m} \\
  \cmidrule(lr){1-8}
  \multirow{2}{*}{a-\ch{SiO2} $\mid$ RSD}
    & $T_{1\text{k}}$ $\downarrow$       & 26.50\std{0.39}m & 28.65\std{0.54}m & 32.27\std{0.61}m & \second{23.76\std{0.32}m} & \first{23.25\std{0.40}m} & 24.08\std{0.65}m \\
    & $T_{1\text{k}}^{Q=0}$ $\downarrow$ & 10.77\std{1.32}h & 13.64\std{2.08}h & 11.20\std{1.42}h & \second{5.82\std{0.77}h} & 6.05\std{0.67}h & \first{24.08\std{0.65}m} \\
  \cmidrule(lr){1-8}
  \multirow{2}{*}{MEG $\mid$ $E$+$C_\text{Li}$}
    & $T_{1\text{k}}$ $\downarrow$       & 53.10\std{0.91}m & 57.38\std{0.82}m & 64.23\std{1.12}m & 47.82\std{0.65}m & \first{46.82\std{0.78}m} & \second{47.44\std{0.88}m} \\
    & $T_{1\text{k}}^{Q=0}$ $\downarrow$ & 88.50\std{17.70}h & 119.54\std{29.89}h & 97.32\std{26.60}h & 56.93\std{12.22}h & \second{48.77\std{12.22}h} & \first{47.44\std{0.88}m} \\
  \cmidrule(lr){1-8}
  \multirow{2}{*}{MEG $\mid$ $G$}
    & $T_{1\text{k}}$ $\downarrow$       & 52.87\std{0.87}m & 57.15\std{0.78}m & 63.96\std{1.08}m & 47.62\std{0.63}m & \first{46.62\std{0.74}m} & \second{47.13\std{0.86}m} \\
    & $T_{1\text{k}}^{Q=0}$ $\downarrow$ & 67.78\std{15.64}h & 95.25\std{19.05}h & 76.14\std{16.37}h & 44.09\std{9.82}h & \second{38.85\std{7.79}h} & \first{47.13\std{0.86}m} \\
  \bottomrule
\end{tabular}
}
\end{table}

\subsection{Model Analysis}
\label{sec:model-analysis}
We provide insights into the effectiveness of design choices and hyperparameter settings in \archname{} with the following model analysis experiments.

\paragraph{Effectiveness of components.}
We evaluate the contribution of each component by cumulatively adding them to a base flow matching model on the MEG $\mid$ $E$+$C_\text{Li}$ configuration.
Starting from \textit{Base}, a standard flow matching model without any charge balance components (\Secref{sec:fm}), we progressively add data-centered conditional element noise (\Secref{sec:ot-noise}) as \textit{+Noise}, then per-step soft linearized projection (\Secref{sec:pcfm}) as \textit{+Proj.}, and finally the discrete projection (\Secref{sec:dp}) to arrive at full \textit{\archname{}}.
Table~\ref{tab:ablation} reports the results. Figure~\ref{fig:ablation} further visualizes the effect of each component, where (a) Base, (b) +Noise, and (c) +Proj.\ correspond to the cumulative addition of components.
Adding the data-centered conditional element noise (+Noise) reduces $|\bar{Q}|$ substantially by giving the generation a starting point closer to charge balance, but without active correction during generation, the charge spread across samples remains large.
The soft linearized projection (+Proj.) addresses this by steering elements toward charge balance at each generation step, as visible in the progressively tighter convergence in Figures~\ref{fig:convergence-base}--\ref{fig:convergence-proj}.
Notably, +Proj.\ also improves inverse design metrics, suggesting that steering element assignments toward charge balance also aligns them more closely with the training data distribution, where all samples are charge balanced.
The final discrete projection brings $P(Q{=}0)$ to 100\% with negligible change in inverse design metrics, confirming that the residual charge after +Proj.\ is small enough to be resolved with minimal element swaps (Figure~\ref{fig:charge-dist}).

We further study the sensitivity of two critical hyperparameters in these components, the element noise scale $\sigma$ and softmax temperature $\tau$, in Appendix~\ref{apx:hp-sensitivity}.

\begin{table}[t]
  \caption{Ablation study on the MEG $\mid$ $E$+$C_\text{Li}$ configuration. \first{Bold} and \second{underline} indicate the best and second best results, respectively.}
  \label{tab:ablation}
  \centering
\begin{tabular}{ll cccc}
  \toprule
  Category & Metric & Base & +Noise & +Proj. & \archname{} \\
  \midrule
  \multirow{6}{*}{\shortstack{Inverse\\Design}}
    & $E$ MAE $\downarrow$           & 10.68\std{0.30} & 10.42\std{0.31} & \second{9.95\std{0.27}} & \first{9.92\std{0.26}} \\
    & $E$ RMSE $\downarrow$          & 13.02\std{0.35} & 12.72\std{0.36} & \first{11.80\std{0.30}} & \second{11.82\std{0.29}} \\
    & $E$ MAPE $\downarrow$          & 21.6\std{0.6}\% & 21.1\std{0.6}\% & \second{20.2\std{0.5}\%} & \first{20.1\std{0.5}\%} \\
    & $C_\text{Li}$ MAE $\downarrow$  & 1.60\std{0.04} & 1.56\std{0.05} & \second{1.51\std{0.04}} & \first{1.49\std{0.04}} \\
    & $C_\text{Li}$ RMSE $\downarrow$ & 1.95\std{0.05} & 1.91\std{0.06} & \second{1.83\std{0.05}} & \first{1.82\std{0.04}} \\
    & $C_\text{Li}$ MAPE $\downarrow$ & 10.7\std{0.4}\% & 10.4\std{0.4}\% & \second{9.9\std{0.3}\%} & \first{9.8\std{0.3}\%} \\
  \midrule
  \multirow{3}{*}{\shortstack{Charge\\Balance}}
    & $P(Q{=}0)$ $\uparrow$  & 1.4\std{0.4}\% & 2.1\std{0.6}\% & \second{16.2\std{2.5}\%} & \first{100.0\std{0.0}\%} \\
    & $|\bar{Q}|$ $\downarrow$  & 34.62\std{5.15} & 13.57\std{2.08} & \second{0.57\std{0.09}} & \first{0.00\std{0.00}} \\
    & $\sigma_Q$ $\downarrow$ & 61.62\std{5.84} & 80.40\std{7.23} & \second{2.41\std{0.38}} & \first{0.00\std{0.00}} \\
  \midrule
  \multirow{2}{*}{Cost}
    & $T_{1\text{k}}$ $\downarrow$      & \first{47.02\std{0.67}m} & \first{47.02\std{0.67}m} & \second{47.19\std{0.68}m} & 47.44\std{0.88}m \\
    & $T_{1\text{k}}^{Q=0}$ $\downarrow$ & 55.98\std{16.01}h & 37.32\std{10.68}h & \second{4.85\std{0.75}h} & \first{47.44\std{0.88}m} \\
  \bottomrule
\end{tabular}

\end{table}

\begin{table}[t]
  \caption{Contribution of components beyond the final discrete projection (DP) on MEG $\mid$ $E$+$C_\text{Li}$. $n_\text{DP}$: mean number of element assignments changed by DP per sample. JSD: Jensen-Shannon divergence of partial radial distribution functions. CN MAE: MAE of first-shell coordination numbers (Appendix~\ref{apx:dp-effect}). \first{Bold} and \second{underline} indicate the best and second best results.}
  \label{tab:dp-ablation}
  \centering
\begin{tabular}{l ccccc}
  \toprule
  Metric & Base & Base+DP & +Noise+DP & +Proj. & \archname{} \\
  \midrule
  $E$ MAPE $\downarrow$ & 21.6\% & 23.9\% & 22.5\% & \second{20.2\%} & \first{20.1\%} \\
  $C_\text{Li}$ MAPE $\downarrow$ & 10.7\% & 11.7\% & 11.0\% & \second{9.9\%} & \first{9.8\%} \\
  $P(Q{=}0)$ $\uparrow$ & 1.4\% & \first{100.0\%} & \first{100.0\%} & 16.2\% & \first{100.0\%} \\
  $n_\text{DP}$ $\downarrow$ & -- & 14.73 & \second{9.21} & -- & \first{0.81} \\
  JSD $\downarrow$ & 0.01475 & 0.01578 & 0.01527 & \first{0.01415} & \second{0.01418} \\
  CN MAE $\downarrow$ & 0.04594 & 0.04899 & 0.04769 & \first{0.04471} & \second{0.04482} \\
  \bottomrule
\end{tabular}

\end{table}

\paragraph{Contribution of components beyond the final discrete projection.}
Since the final discrete projection (DP) alone guarantees $P(Q{=}0) = 100\%$, we also apply it directly to Base and +Noise, denoted \textit{Base+DP} and \textit{+Noise+DP} (Table~\ref{tab:dp-ablation}).
Base+DP changes 14.73 element assignments per sample and degrades the inverse design and structural metrics relative to Base.
The data-centered conditional element noise and the soft linearized projection reduce this number to 9.21 and 0.81, while \archname{} matches the metrics of +Proj.
Thus, DP guarantees exact charge balance, while the preceding components reduce the charge residual enough for DP to preserve these metrics.

\paragraph{Effect of the final discrete projection.}
Across the four configurations, DP changes 0.59 to 0.81 element assignments per sample on average, and at most four in any sample (Appendix~\ref{apx:dp-effect}).
With atom positions unchanged, DP changes the structural metrics by at most 0.25\%, and on MEG $\mid$ $E$+$C_\text{Li}$ the mean atomic percentage of each element by at most 0.01 percentage points.

\paragraph{Computational overhead of components.}
As shown in Table~\ref{tab:ablation}, the data-centered conditional element noise adds no generation time since it only affects the initial noise sampling, and the soft linearized projection and DP together add less than 1\% to $T_{1\text{k}}$.
Meanwhile, $T_{1\text{k}}^{Q=0}$ drops from 55.98 hours to 47.44 minutes as each component improves $P(Q{=}0)$.

\section{Conclusion}
This work introduces \archname{}, a generative inverse design method for amorphous materials that guarantees the generation of formally charge balanced samples.
Built on flow matching, \archname{} combines a data-centered conditional element noise that starts the generation closer to charge balance, a per-step soft linearized projection that steers element assignments toward charge balance via a relaxed constraint, and a final discrete projection that guarantees exact charge balance through minimum-cost element swaps.
Experiments on two amorphous material datasets provide evidence that \archname{} achieves its design goals without compromising inverse design performance or adding significant computational overhead.
Broader impacts and limitations are discussed in Appendix~\ref{apx:discussion}.

\begin{ack}
We thank Jonas A. Finkler (Aalborg University) and Tao Du (The Hong Kong Polytechnic University) for their contributions to datasets and implementation code in our prior joint works~\citep{finkler2025inverse,lin2026amshortcut} that are partially reused in this work, and, together with Suresh Bishnoi (Aalborg University), for their involvement in the preliminary discussions leading to this work.

This work was supported by a research grant (VIL57373) from VILLUM FONDEN.
The authors declare no competing interests.
\end{ack}

\bibliographystyle{plainnat}
\bibliography{references}

\appendix

\section{Charge Balance Component Hyperparameter Sensitivity}
\label{apx:hp-sensitivity}

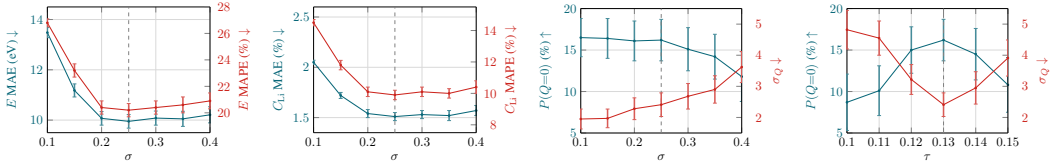
\begin{figure}[ht]
  \centering
  \begin{subfigure}[b]{0.24\textwidth}
    \centering
    \resizebox{\textwidth}{!}{%
\begin{tikzpicture}
\pgfplotsset{
  every axis/.append style={
    width=6cm,
    height=5.0cm,
    xmin=0.10, xmax=0.40,
    xtick={0.10,0.20,0.30,0.40},
  },
}
\begin{axis}[
  axis y line*=left,
  xlabel={$\sigma$},
  ylabel={$E$ MAE (eV) $\downarrow$},
  ylabel style={color=plotblue},
  yticklabel style={color=plotblue, text width=1.2em, align=right},
  ymin=9.5, ymax=14.5,
  grid=major,
  grid style={gray!30},
]
\addplot[plotblue, thick, mark=*, mark size=0.6pt,
  error bars/.cd, y dir=both, y explicit,
  error bar style={plotblue!75, line width=1.2pt}, error mark options={rotate=90, mark size=1.5pt, plotblue!75}]
  coordinates {
    (0.10, 13.48) +- (0, 0.20)
    (0.15, 11.18) +- (0, 0.26)
    (0.20, 10.07) +- (0, 0.27)
    (0.25, 9.95) +- (0, 0.27)
    (0.30, 10.08) +- (0, 0.28)
    (0.35, 10.05) +- (0, 0.30)
    (0.40, 10.21) +- (0, 0.32)
  };
\draw[dashed, gray, thin] (axis cs:0.25, 9.5) -- (axis cs:0.25, 14.5);
\end{axis}
\begin{axis}[
  axis y line*=right,
  axis x line=none,
  ylabel={$E$ MAPE (\%) $\downarrow$},
  ylabel style={color=plotred},
  yticklabel style={color=plotred, text width=1.2em, align=right},
  ymin=18.5, ymax=28.0,
]
\addplot[plotred, thick, mark=*, mark size=0.6pt,
  error bars/.cd, y dir=both, y explicit,
  error bar style={plotred!75, line width=1.2pt}, error mark options={rotate=90, mark size=1.5pt, plotred!75}]
  coordinates {
    (0.10, 26.8) +- (0, 0.3)
    (0.15, 23.2) +- (0, 0.5)
    (0.20, 20.4) +- (0, 0.5)
    (0.25, 20.2) +- (0, 0.5)
    (0.30, 20.4) +- (0, 0.5)
    (0.35, 20.6) +- (0, 0.6)
    (0.40, 20.9) +- (0, 0.6)
  };
\end{axis}
\end{tikzpicture}%
}
    \vspace{-10pt}
    \caption{$E$ MAE, MAPE vs.\ $\sigma$}
    \label{fig:hp-sigma-e}
  \end{subfigure}
  \hfill
  \begin{subfigure}[b]{0.24\textwidth}
    \centering
    \resizebox{\textwidth}{!}{%
\begin{tikzpicture}
\pgfplotsset{
  every axis/.append style={
    width=6cm,
    height=5.0cm,
    xmin=0.10, xmax=0.40,
    xtick={0.10,0.20,0.30,0.40},
  },
}
\begin{axis}[
  axis y line*=left,
  xlabel={$\sigma$},
  ylabel={$C_\text{Li}$ MAE (\%) $\downarrow$},
  ylabel style={color=plotblue},
  yticklabel style={color=plotblue, text width=1.2em, align=right},
  ymin=1.35, ymax=2.60,
  grid=major,
  grid style={gray!30},
]
\addplot[plotblue, thick, mark=*, mark size=0.6pt,
  error bars/.cd, y dir=both, y explicit,
  error bar style={plotblue!75, line width=1.2pt}, error mark options={rotate=90, mark size=1.5pt, plotblue!75}]
  coordinates {
    (0.10, 2.05) +- (0, 0.02)
    (0.15, 1.72) +- (0, 0.03)
    (0.20, 1.54) +- (0, 0.04)
    (0.25, 1.51) +- (0, 0.04)
    (0.30, 1.53) +- (0, 0.04)
    (0.35, 1.52) +- (0, 0.05)
    (0.40, 1.57) +- (0, 0.05)
  };
\draw[dashed, gray, thin] (axis cs:0.25, 1.35) -- (axis cs:0.25, 2.60);
\end{axis}
\begin{axis}[
  axis y line*=right,
  axis x line=none,
  ylabel={$C_\text{Li}$ MAPE (\%) $\downarrow$},
  ylabel style={color=plotred},
  yticklabel style={color=plotred, text width=1.2em, align=right},
  ymin=7.5, ymax=15.5,
]
\addplot[plotred, thick, mark=*, mark size=0.6pt,
  error bars/.cd, y dir=both, y explicit,
  error bar style={plotred!75, line width=1.2pt}, error mark options={rotate=90, mark size=1.5pt, plotred!75}]
  coordinates {
    (0.10, 14.5) +- (0, 0.2)
    (0.15, 11.8) +- (0, 0.3)
    (0.20, 10.1) +- (0, 0.3)
    (0.25, 9.9) +- (0, 0.3)
    (0.30, 10.1) +- (0, 0.3)
    (0.35, 10.0) +- (0, 0.3)
    (0.40, 10.4) +- (0, 0.4)
  };
\end{axis}
\end{tikzpicture}%
}
    \vspace{-10pt}
    \caption{$C_\text{Li}$ MAE, MAPE vs.\ $\sigma$}
    \label{fig:hp-sigma-cli}
  \end{subfigure}
  \hfill
  \begin{subfigure}[b]{0.24\textwidth}
    \centering
    \resizebox{\textwidth}{!}{%
\begin{tikzpicture}
\pgfplotsset{
  every axis/.append style={
    width=6cm,
    height=5.0cm,
    xmin=0.10, xmax=0.40,
    xtick={0.10,0.20,0.30,0.40},
  },
}
\begin{axis}[
  axis y line*=left,
  xlabel={$\sigma$},
  ylabel={$P(Q{=}0)$ (\%) $\uparrow$},
  ylabel style={color=plotblue},
  yticklabel style={color=plotblue, text width=1.2em, align=right},
  ymin=5, ymax=20,
  grid=major,
  grid style={gray!30},
]
\addplot[plotblue, thick, mark=*, mark size=0.6pt,
  error bars/.cd, y dir=both, y explicit,
  error bar style={plotblue!75, line width=1.2pt}, error mark options={rotate=90, mark size=1.5pt, plotblue!75}]
  coordinates {
    (0.10, 16.5) +- (0, 2.3)
    (0.15, 16.4) +- (0, 2.4)
    (0.20, 16.1) +- (0, 2.4)
    (0.25, 16.2) +- (0, 2.5)
    (0.30, 15.1) +- (0, 2.6)
    (0.35, 14.2) +- (0, 2.7)
    (0.40, 11.8) +- (0, 3.0)
  };
\draw[dashed, gray, thin] (axis cs:0.25, 5) -- (axis cs:0.25, 20);
\end{axis}
\begin{axis}[
  axis y line*=right,
  axis x line=none,
  ylabel={$\sigma_Q$ $\downarrow$},
  ylabel style={color=plotred},
  yticklabel style={color=plotred, text width=1.2em, align=right},
  ymin=1.5, ymax=5.5,
]
\addplot[plotred, thick, mark=*, mark size=0.6pt,
  error bars/.cd, y dir=both, y explicit,
  error bar style={plotred!75, line width=1.2pt}, error mark options={rotate=90, mark size=1.5pt, plotred!75}]
  coordinates {
    (0.10, 1.95) +- (0, 0.32)
    (0.15, 1.97) +- (0, 0.30)
    (0.20, 2.28) +- (0, 0.35)
    (0.25, 2.41) +- (0, 0.38)
    (0.30, 2.68) +- (0, 0.41)
    (0.35, 2.90) +- (0, 0.44)
    (0.40, 3.62) +- (0, 0.50)
  };
\end{axis}
\end{tikzpicture}%
}
    \vspace{-10pt}
    \caption{$P(Q{=}0)$, $\sigma_Q$ vs.\ $\sigma$}
    \label{fig:hp-sigma-q}
  \end{subfigure}
  \hfill
  \begin{subfigure}[b]{0.24\textwidth}
    \centering
    \resizebox{\textwidth}{!}{%
\begin{tikzpicture}
\pgfplotsset{
  every axis/.append style={
    width=6cm,
    height=5.0cm,
    xmin=0.10, xmax=0.15,
    xtick={0.10,0.11,0.12,0.13,0.14,0.15},
  },
}
\begin{axis}[
  axis y line*=left,
  xlabel={$\tau$},
  ylabel={$P(Q{=}0)$ (\%) $\uparrow$},
  ylabel style={color=plotblue},
  yticklabel style={color=plotblue, text width=1.2em, align=right},
  ymin=5, ymax=20,
  grid=major,
  grid style={gray!30},
]
\addplot[plotblue, thick, mark=*, mark size=0.6pt,
  error bars/.cd, y dir=both, y explicit,
  error bar style={plotblue!75, line width=1.2pt}, error mark options={rotate=90, mark size=1.5pt, plotblue!75}]
  coordinates {
    (0.10, 8.7) +- (0, 3.4)
    (0.11, 10.1) +- (0, 3.0)
    (0.12, 15.0) +- (0, 2.8)
    (0.13, 16.2) +- (0, 2.5)
    (0.14, 14.5) +- (0, 3.1)
    (0.15, 10.8) +- (0, 3.3)
  };
\draw[dashed, gray, thin] (axis cs:0.13, 5) -- (axis cs:0.13, 20);
\end{axis}
\begin{axis}[
  axis y line*=right,
  axis x line=none,
  ylabel={$\sigma_Q$ $\downarrow$},
  ylabel style={color=plotred},
  yticklabel style={color=plotred, text width=1.2em, align=right},
  ymin=1.5, ymax=5.5,
]
\addplot[plotred, thick, mark=*, mark size=0.6pt,
  error bars/.cd, y dir=both, y explicit,
  error bar style={plotred!75, line width=1.2pt}, error mark options={rotate=90, mark size=1.5pt, plotred!75}]
  coordinates {
    (0.10, 4.82) +- (0, 0.62)
    (0.11, 4.55) +- (0, 0.55)
    (0.12, 3.22) +- (0, 0.48)
    (0.13, 2.41) +- (0, 0.38)
    (0.14, 2.95) +- (0, 0.52)
    (0.15, 3.91) +- (0, 0.58)
  };
\end{axis}
\end{tikzpicture}%
}
    \vspace{-10pt}
    \caption{$P(Q{=}0)$, $\sigma_Q$ vs.\ $\tau$}
    \label{fig:hp-tau-q}
  \end{subfigure}
  \caption{Hyperparameter sensitivity on MEG $\mid$ $E$+$C_\text{Li}$.}
  \label{fig:hp-sensitivity}
\end{figure}

We study the sensitivity of two critical hyperparameters introduced by the charge balance components on the MEG $\mid$ $E$+$C_\text{Li}$ configuration: the element noise scale $\sigma$ of the data-centered conditional element noise (\Secref{sec:ot-noise}) and the softmax temperature $\tau$ of the soft linearized projection (\Secref{sec:pcfm}).
We report $P(Q{=}0)$ and $\sigma_Q$ on the +Proj.\ variant, since with the final discrete projection, full \archname{} always achieves $P(Q{=}0) = 100\%$, which would mask the effect of these hyperparameters.
Figure~\ref{fig:hp-sensitivity} summarizes the results.

For $\sigma$ (Figures~\ref{fig:hp-sigma-e}--\ref{fig:hp-sigma-q}), when it is too small, the element noise (\Eqref{eq:ot-noise}) concentrates tightly around the one-hot vectors, reducing the diversity of flow paths and degrading inverse design MAE and MAPE.
Increasing $\sigma$ to 0.25 steadily improves both $E$ and $C_\text{Li}$ MAE and MAPE.
Beyond 0.25, the noise overwhelms the one-hot centering and approaches the isotropic Gaussian of standard flow-matching, causing $P(Q{=}0)$ to drop and $\sigma_Q$ to rise, while degrading MAE and MAPE slightly.

For $\tau$ (Figure~\ref{fig:hp-tau-q}), when it is too small, $\softmax(\hat{\mE}_1/\tau)$ saturates and its gradient vanishes, making the soft linearized projection (\Eqref{eq:gn-projection}) unstable; when it is too large, $Q_\text{soft}$ deviates from the discrete charge $Q$, reducing projection accuracy.
The optimum $\tau{=}0.13$ balances both effects.

\section{EGNN Architecture}
\label{apx:egnn}

Given an input sample $\gM=(\mL, \mX, \mE)$, the input graph $\gG=(\gV, \gE)$ to the EGNN is composed of atoms in the sample as nodes in the node set $\gV$, and each edge in the edge set $\gE$ connects a pair of atoms with distances less than a cutoff radius $r_\text{cut}$. The distances are computed with periodic boundary conditions.

The EGNN is composed of $L$ equivariant layers.
The $l$-th layer takes as input:
1) node features $\mH^{(l)} \in \sR^{n_a \times d_h}$ containing information about the corresponding atoms at the $l$-th layer;
2) positional coordinates $\mX^{(l)} \in \sR^{n_a \times k \times 3}$ of the atoms, where $k$ is the number of vector channels~\citep{levy2023using}; and
3) edge set $\gE$ of the graph $\gG$, with edge attributes $\ve_{ij}$ assigned to each edge.

For the initial layer, the node features $\mH_i^{(0)}$ are assembled by concatenating:
1) the time step $t$;
2) the element embedding $\mE_i \in \sR^{d_E}$; and
3) a linear projection of the target property vector $\vy$.
The positions $\mX^{(0)}$ are the original positions $\mX$ replicated across $k$ channels.
The edge attributes $\ve_{ij}$ are derived from the distance embedding:
\begin{equation}
\ve_{ij} = \tanh\left(\frac{\|\mX_i - \mX_j - \mathbf{o}_{ij}\|^2}{r_{\text{cut}}^2}\right) \cdot 2 - 1
\end{equation}
where $\mathbf{o}_{ij}$ is the offset vector accounting for periodic boundary conditions.

Each layer updates the node features and positional coordinates, incorporating self-attention~\citep{DBLP:conf/nips/VaswaniSPUJGKP17}:
\begin{equation}
  \begin{aligned}
    \mathbf{m}_{ij}^{(l)} &= \phi_e^{(l)}(\mH_i^{(l-1)}, \mH_j^{(l-1)}, \ve_{ij}) \\
    \alpha_{ij}^{(l)} &= \sigma(\text{MLP}_{\text{att}}(\mathbf{m}_{ij}^{(l)})) \\
    \hat{\mathbf{m}}_{ij}^{(l)} &= \alpha_{ij}^{(l)} \cdot \mathbf{m}_{ij}^{(l)} \\
    \mH_i^{(l)} &= \mH_i^{(l-1)} + \phi_H^{(l)}\left(\mH_i^{(l-1)}, \sum_{j \in N(i)} \frac{f_\text{cut}(d_{ij}^{(0)}) \cdot \hat{\mathbf{m}}_{ij}^{(l)}}{n_\text{norm}}\right) \\
    \mathbf{\Phi}_{ij}^{(l)} &= \text{MLP}_{\text{coord}}([\mH_i^{(l)}, \mH_j^{(l)}, \ve_{ij}]) \in \sR^{k \times k} \\
    \mathbf{d}_{ij}^{(l)} &= \mX_i^{(l-1)} - \mX_j^{(l-1)} - \mathbf{o}_{ij} \\
    \mX_i^{(l)} &= \mX_i^{(l-1)} + \sum_{j \in N(i)} \frac{1}{n_\text{norm}} \cdot \mathbf{\Phi}_{ij}^{(l)} \cdot \mathbf{d}_{ij}^{(l)}
  \end{aligned}
\end{equation}
where $N(i)$ represents the neighbors of atom $i$ derived from the edge set $\gE$,
$\sigma$ is the sigmoid activation function for self-attention,
and $n_\text{norm}$ is a normalization factor to ensure numerical stability.
$\phi_e^{(l)}$ and $\phi_H^{(l)}$ are multi-layer perceptrons (MLPs) with SiLU activation functions and layer normalization, structured as:
\begin{equation}
  \begin{aligned}
    \phi_e^{(l)}(\mH_i, \mH_j, \ve_{ij}) &= \text{MLP}_{\text{edge}}([\mH_i, \mH_j, \ve_{ij}]) \\
    \phi_H^{(l)}(\mH_i, \mathbf{m}_{\text{agg}}) &= \text{MLP}_{\text{node}}([\mH_i, \mathbf{m}_{\text{agg}}])
  \end{aligned}
\end{equation}
$\mathbf{\Phi}_{ij}^{(l)}$ is a learned transformation matrix that maps between the $k$ vector channels.

A smooth cutoff function is used to prevent discontinuities when atoms leave or enter the cutoff radius:
\begin{equation}
  f_\text{cut}(r) = 2 \tanh\left(1 - \frac{\min(r, r_\text{cut})}{r_\text{cut}}\right)^2
\end{equation}

At the last layer, the EGNN outputs $\mH^{(L)}$ and $\mX^{(L)}$ as the final node features and positional coordinates, respectively. The element velocity $\vv_\mE$ is taken directly from $\mH^{(L)}$, and the position velocity $\vv_\mX$ is computed as the deviation between the original positions and the first channel of the output positions, scaled by a learnable scalar parameter.

\begin{figure}[ht]
  \centering
  \begin{subfigure}[b]{0.24\textwidth}
    \centering
    \resizebox{\textwidth}{!}{%
\begin{tikzpicture}
\pgfplotsset{
  every axis/.append style={
    width=6cm,
    height=5.0cm,
    xmin=2, xmax=6,
    xtick={2,3,4,5,6},
  },
}
\begin{axis}[
  axis y line*=left,
  xlabel={$L$},
  ylabel={$E$ MAPE (\%) $\downarrow$},
  ylabel style={color=plotblue},
  yticklabel style={color=plotblue, text width=1.2em, align=right},
  ymin=16, ymax=32,
  ytick={16,20,24,28,32},
  grid=major,
  grid style={gray!30},
]
\addplot[plotblue, thick, mark=*, mark size=0.6pt,
  error bars/.cd, y dir=both, y explicit,
  error bar style={plotblue!75, line width=1.2pt}, error mark options={rotate=90, mark size=1.5pt, plotblue!75}]
  coordinates {
    (2, 28.4) +- (0, 1.5)
    (3, 23.1) +- (0, 0.8)
    (4, 20.5) +- (0, 0.6)
    (5, 21.0) +- (0, 0.8)
    (6, 23.2) +- (0, 1.1)
  };
\draw[dashed, gray, thin] (axis cs:4, 16) -- (axis cs:4, 32);
\end{axis}
\begin{axis}[
  axis y line*=right,
  axis x line=none,
  ylabel={$C_\text{Li}$ MAPE (\%) $\downarrow$},
  ylabel style={color=plotred},
  yticklabel style={color=plotred, text width=1.2em, align=right},
  ymin=8, ymax=20,
  ytick={8,12,16,20},
]
\addplot[plotred, thick, mark=*, mark size=0.6pt,
  error bars/.cd, y dir=both, y explicit,
  error bar style={plotred!75, line width=1.2pt}, error mark options={rotate=90, mark size=1.5pt, plotred!75}]
  coordinates {
    (2, 16.7) +- (0, 1.1)
    (3, 12.3) +- (0, 0.6)
    (4, 10.3) +- (0, 0.4)
    (5, 10.9) +- (0, 0.5)
    (6, 12.1) +- (0, 0.7)
  };
\end{axis}
\end{tikzpicture}%
}
    \vspace{-10pt}
    \caption{MAPE vs.\ $L$}
    \label{fig:hp-L}
  \end{subfigure}
  \hfill
  \begin{subfigure}[b]{0.24\textwidth}
    \centering
    \resizebox{\textwidth}{!}{%
\begin{tikzpicture}
\pgfplotsset{
  every axis/.append style={
    width=6cm,
    height=5.0cm,
    xmin=64, xmax=256,
    xtick={64,128,192,256},
  },
}
\begin{axis}[
  axis y line*=left,
  xlabel={$d_h$},
  ylabel={$E$ MAPE (\%) $\downarrow$},
  ylabel style={color=plotblue},
  yticklabel style={color=plotblue, text width=1.2em, align=right},
  ymin=16, ymax=32,
  ytick={16,20,24,28,32},
  grid=major,
  grid style={gray!30},
]
\addplot[plotblue, thick, mark=*, mark size=0.6pt,
  error bars/.cd, y dir=both, y explicit,
  error bar style={plotblue!75, line width=1.2pt}, error mark options={rotate=90, mark size=1.5pt, plotblue!75}]
  coordinates {
    (64, 22.8) +- (0, 0.8)
    (128, 20.5) +- (0, 0.6)
    (192, 19.9) +- (0, 0.6)
    (256, 19.5) +- (0, 0.7)
  };
\draw[dashed, gray, thin] (axis cs:128, 16) -- (axis cs:128, 32);
\end{axis}
\begin{axis}[
  axis y line*=right,
  axis x line=none,
  ylabel={$C_\text{Li}$ MAPE (\%) $\downarrow$},
  ylabel style={color=plotred},
  yticklabel style={color=plotred, text width=1.2em, align=right},
  ymin=8, ymax=20,
  ytick={8,12,16,20},
]
\addplot[plotred, thick, mark=*, mark size=0.6pt,
  error bars/.cd, y dir=both, y explicit,
  error bar style={plotred!75, line width=1.2pt}, error mark options={rotate=90, mark size=1.5pt, plotred!75}]
  coordinates {
    (64, 12.7) +- (0, 0.6)
    (128, 10.3) +- (0, 0.4)
    (192, 9.8) +- (0, 0.4)
    (256, 10.0) +- (0, 0.4)
  };
\end{axis}
\end{tikzpicture}%
}
    \vspace{-10pt}
    \caption{MAPE vs.\ $d_h$}
    \label{fig:hp-dh}
  \end{subfigure}
  \hfill
  \begin{subfigure}[b]{0.24\textwidth}
    \centering
    \resizebox{\textwidth}{!}{%
\begin{tikzpicture}
\pgfplotsset{
  every axis/.append style={
    width=6cm,
    height=5.0cm,
    xmin=4, xmax=12,
    xtick={4,6,8,10,12},
  },
}
\begin{axis}[
  axis y line*=left,
  xlabel={$k$},
  ylabel={$E$ MAPE (\%) $\downarrow$},
  ylabel style={color=plotblue},
  yticklabel style={color=plotblue, text width=1.2em, align=right},
  ymin=16, ymax=32,
  ytick={16,20,24,28,32},
  grid=major,
  grid style={gray!30},
]
\addplot[plotblue, thick, mark=*, mark size=0.6pt,
  error bars/.cd, y dir=both, y explicit,
  error bar style={plotblue!75, line width=1.2pt}, error mark options={rotate=90, mark size=1.5pt, plotblue!75}]
  coordinates {
    (4, 23.8) +- (0, 1.0)
    (6, 21.2) +- (0, 0.7)
    (8, 20.5) +- (0, 0.5)
    (10, 20.3) +- (0, 0.6)
    (12, 20.6) +- (0, 0.5)
  };
\draw[dashed, gray, thin] (axis cs:8, 16) -- (axis cs:8, 32);
\end{axis}
\begin{axis}[
  axis y line*=right,
  axis x line=none,
  ylabel={$C_\text{Li}$ MAPE (\%) $\downarrow$},
  ylabel style={color=plotred},
  yticklabel style={color=plotred, text width=1.2em, align=right},
  ymin=8, ymax=20,
  ytick={8,12,16,20},
]
\addplot[plotred, thick, mark=*, mark size=0.6pt,
  error bars/.cd, y dir=both, y explicit,
  error bar style={plotred!75, line width=1.2pt}, error mark options={rotate=90, mark size=1.5pt, plotred!75}]
  coordinates {
    (4, 12.6) +- (0, 0.6)
    (6, 10.8) +- (0, 0.4)
    (8, 9.8) +- (0, 0.3)
    (10, 9.7) +- (0, 0.4)
    (12, 9.6) +- (0, 0.3)
  };
\end{axis}
\end{tikzpicture}%
}
    \vspace{-10pt}
    \caption{MAPE vs.\ $k$}
    \label{fig:hp-k}
  \end{subfigure}
  \hfill
  \begin{subfigure}[b]{0.24\textwidth}
    \centering
    \resizebox{\textwidth}{!}{%
\begin{tikzpicture}
\pgfplotsset{
  every axis/.append style={
    width=6cm,
    height=5.0cm,
    xmin=5.5, xmax=7.5,
    xtick={5.5,6.0,6.5,7.0,7.5},
  },
}
\begin{axis}[
  axis y line*=left,
  xlabel={$r_\text{cut}$ (\AA)},
  ylabel={$E$ MAPE (\%) $\downarrow$},
  ylabel style={color=plotblue},
  yticklabel style={color=plotblue, text width=1.2em, align=right},
  ymin=16, ymax=32,
  ytick={16,20,24,28,32},
  grid=major,
  grid style={gray!30},
]
\addplot[plotblue, thick, mark=*, mark size=0.6pt,
  error bars/.cd, y dir=both, y explicit,
  error bar style={plotblue!75, line width=1.2pt}, error mark options={rotate=90, mark size=1.5pt, plotblue!75}]
  coordinates {
    (5.5, 27.1) +- (0, 1.4)
    (6.0, 22.5) +- (0, 0.8)
    (6.5, 20.5) +- (0, 0.6)
    (7.0, 20.8) +- (0, 0.6)
    (7.5, 21.5) +- (0, 0.7)
  };
\draw[dashed, gray, thin] (axis cs:6.5, 16) -- (axis cs:6.5, 32);
\end{axis}
\begin{axis}[
  axis y line*=right,
  axis x line=none,
  ylabel={$C_\text{Li}$ MAPE (\%) $\downarrow$},
  ylabel style={color=plotred},
  yticklabel style={color=plotred, text width=1.2em, align=right},
  ymin=8, ymax=20,
  ytick={8,12,16,20},
]
\addplot[plotred, thick, mark=*, mark size=0.6pt,
  error bars/.cd, y dir=both, y explicit,
  error bar style={plotred!75, line width=1.2pt}, error mark options={rotate=90, mark size=1.5pt, plotred!75}]
  coordinates {
    (5.5, 16.2) +- (0, 0.9)
    (6.0, 12.3) +- (0, 0.5)
    (6.5, 10.3) +- (0, 0.3)
    (7.0, 10.5) +- (0, 0.4)
    (7.5, 11.1) +- (0, 0.5)
  };
\end{axis}
\end{tikzpicture}%
}
    \vspace{-10pt}
    \caption{MAPE vs.\ $r_\text{cut}$}
    \label{fig:hp-rcut}
  \end{subfigure}
  \caption{EGNN architectural hyperparameter sensitivity on the MEG $\mid$ $E$+$C_\text{Li}$ configuration. Dashed lines mark the chosen values.}
  \label{fig:arch-sensitivity}
\end{figure}
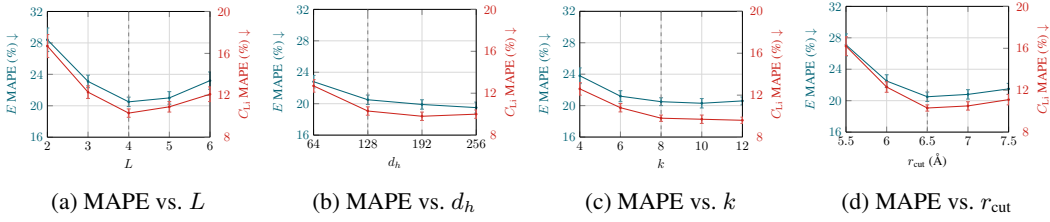

\paragraph{Parameter choices.}
We use $L=4$ EGNN layers with hidden dimension $d_h=128$, $k=8$ vector channels, cutoff radius $r_\text{cut}=6.5$\,\AA, and normalization factor $n_\text{norm}=40$. The self-attention hidden dimension is 128.
Among these, $L$, $d_h$, and $k$ all govern model capacity. Increasing any of them beyond the optimum yields marginal accuracy improvement while reducing computational efficiency.
$L$ additionally controls the receptive depth over the material graph. Too few layers limit long-range information propagation, while too many risk over-smoothing.
The cutoff radius $r_\text{cut}=6.5$\,\AA\ is chosen to cover typical nearest- and second-nearest-neighbor shells in the amorphous structures, and $n_\text{norm}=40$ is set to the average number of neighbors within $r_\text{cut}$ to stabilize the aggregation magnitudes.
Figure~\ref{fig:arch-sensitivity} empirically confirms these choices.

\paragraph{Parameter sensitivity.}
Figure~\ref{fig:arch-sensitivity} shows the effect of each architectural hyperparameter on $E$ and $C_\text{Li}$ MAPE, evaluated on the MEG $\mid$ $E$+$C_\text{Li}$ configuration.
$L$ exhibits a steep improvement from 2 to 4 layers, followed by degradation at deeper settings consistent with over-smoothing.
$d_h$ shows a large jump from 64 to 128 with marginal returns thereafter, thus we select $d_h{=}128$ as the most efficient operating point; $k$ follows a similar pattern, where we select $k{=}8$ as the optimum.
$r_\text{cut}$ displays a clear U-shape centered at 6.5\,\AA: too small a radius misses relevant interactions, while too large a radius introduces noisy long-range edges.

\section{Algorithms}
\label{apx:algorithms}

\subsection{The Full Inference Procedure}
\label{apx:inference-algorithm}

Algorithm~\ref{alg:inference} presents the full inference procedure of \archname{}, combining the data-centered conditional element noise (\Secref{sec:ot-noise}), per-step soft linearized projection (\Secref{sec:pcfm}), and final discrete projection (\Secref{sec:dp}).
At each integration step, the velocity network predicts position and element velocities.
Positions are advanced by a standard Euler step, while elements undergo a clean extrapolation to $t{=}1$ followed by the soft linearized projection that reduces the charge residual.
The projected clean state is then re-interpolated along the flow path to obtain the element state for the next step.
After the final step, element assignments are discretized via $\argmax$ and corrected to exact charge balance by the final discrete projection (\Algref{alg:dp}).

\begin{algorithm}[ht]
\caption{\archname{} inference procedure.}
\label{alg:inference}
\begin{algorithmic}[1]
\REQUIRE Lattice $\mL$, number of atoms $n_a$, target properties $\vy$, number of steps $T$, element frequencies $\{f_k\}_{k=1}^{d_E}$, formal charges $\vc \in \sZ^{d_E}$, noise scale $\sigma$, softmax temperature $\tau$
\ENSURE Generated material $\gM = (\mL, \mX_1, \mE)$ with $Q = 0$
\STATE Sample positions $\mX_0$ uniformly in the unit cell
\STATE Sample element noise: for each atom $i$, draw $k_i \sim \mathrm{Cat}(f_1, \ldots, f_{d_E})$, \, $\mE_{0,i} \sim \gN(\ve_{k_i},\, \sigma^2 \mI)$ \COMMENT{\Eqref{eq:marginal}}
\STATE $\Delta t \leftarrow 1 / T$
\FOR{$i = 0$ \TO $T - 1$}
    \STATE $t \leftarrow i \cdot \Delta t$
    \STATE $(\vv_\mX,\, \vv_\mE) \leftarrow \vv_\theta(\gM_t,\, \vy,\, t)$ \COMMENT{Velocity prediction}
    \STATE $\mX_{t+\Delta t} \leftarrow \mX_t + \Delta t \cdot \vv_\mX$ \COMMENT{Euler step for positions}
    \STATE $\hat{\mE}_1 \leftarrow \mE_t + (1 - t) \cdot \vv_\mE$ \COMMENT{Clean extrapolation (\Eqref{eq:clean-extrap})}
    \STATE $Q \leftarrow \vone^\top \mathrm{onehot}(\argmax(\hat{\mE}_1))\, \vc$ \COMMENT{Discrete charge}
    \IF{$Q \neq 0$}
        \STATE $Q_\text{soft} \leftarrow \vone^\top \softmax(\hat{\mE}_1 / \tau)\, \vc$
        \STATE Compute $\nabla_{\hat{\mE}_1} Q_\text{soft}$ via backpropagation
        \STATE $\hat{\mE}_1^\text{proj} \leftarrow \hat{\mE}_1 - \dfrac{Q}{\|\nabla_{\hat{\mE}_1} Q_\text{soft}\|^2}\,\nabla_{\hat{\mE}_1} Q_\text{soft}$ \COMMENT{\Eqref{eq:gn-projection}}
    \ELSE
        \STATE $\hat{\mE}_1^\text{proj} \leftarrow \hat{\mE}_1$
    \ENDIF
    \STATE $\mE_{t+\Delta t} \leftarrow (1 - t - \Delta t)\,\mE_0 + (t + \Delta t)\,\hat{\mE}_1^\text{proj}$ \COMMENT{\Eqref{eq:ot-interpolant}}
\ENDFOR
\STATE $e_i \leftarrow \argmax_j \hat{\mE}_{1,ij}$ for all $i$ \COMMENT{Discretize elements}
\STATE $\{e_i'\} \leftarrow \text{Final-Discrete-Projection}(\hat{\mE}_1,\, \vc)$ \COMMENT{\Algref{alg:dp}}
\RETURN $\gM = (\mL,\, \mX_1,\, \mathrm{onehot}(\{e_i'\}))$
\end{algorithmic}
\end{algorithm}

\subsection{Final Discrete Projection}
\label{apx:dp-algorithm}

Algorithm~\ref{alg:dp} solves the optimization problem in \Eqref{eq:dp-objective} via dynamic programming.
Since formal charges $c_j$ are integers, the total charge delta from the original assignments can serve as a discrete state variable for the dynamic programming.
The algorithm processes atoms one at a time: at each atom, it considers every possible element assignment and extends the cost table by accounting for the resulting change in charge and the logit gap sacrificed by the swap.
After all $n_a$ atoms are processed, the optimal corrected assignments $\{e_i'\}$ satisfying $\sum_i c_{e_i'} = 0$ are recovered by tracing back through the cost table from the target state.
The overall complexity is $O(n_a \cdot |Q_\text{max}| \cdot d_E)$, where $|Q_\text{max}|$ is the range of reachable charge states.
In our implementation, the inner loop over elements $j \in \{1, \ldots, d_E\}$ is vectorized into matrix operations, computing all swap costs and charge deltas for the $d_E$ candidates simultaneously at each atom $i$ to reduce wall-clock time.

\begin{algorithm}[ht]
\caption{Minimum-cost discrete projection via dynamic programming.}
\label{alg:dp}
\begin{algorithmic}[1]
\REQUIRE Element logits $\hat{\mE}_1 \in \sR^{n_a \times d_E}$ at $t=1$, formal charges $\vc \in \sZ^{d_E}$
\ENSURE Corrected assignments $\{e_i'\}_{i=1}^{n_a}$ with $\sum_i c_{e_i'} = 0$
\STATE $e_i \leftarrow \argmax_j \hat{\mE}_{1,ij}$ for all $i$ \COMMENT{Current assignments}
\STATE $Q \leftarrow \sum_{i=1}^{n_a} c_{e_i}$ \COMMENT{Current total charge}
\IF{$Q = 0$}
    \RETURN $\{e_i\}$ \COMMENT{Already balanced}
\ENDIF
\STATE Initialize cost table: $\mathrm{cost}[0, 0] \leftarrow 0$, all others $\leftarrow \infty$
\FOR{$i = 1$ \TO $n_a$}
    \FOR{each element $j \in \{1, \ldots, d_E\}$}
        \STATE $s \leftarrow \hat{\mE}_{1,i,e_i} - \hat{\mE}_{1,i,j}$ \COMMENT{Swap cost}
        \STATE $\delta \leftarrow c_j - c_{e_i}$ \COMMENT{Charge delta}
        \FOR{each reachable state $q$}
            \STATE $\mathrm{cost}[i, q + \delta] \leftarrow \min(\mathrm{cost}[i, q + \delta],\; \mathrm{cost}[i{-}1, q] + s)$
        \ENDFOR
    \ENDFOR
\ENDFOR
\STATE Backtrack from state $q = -Q$ to recover optimal $\{e_i'\}$
\RETURN $\{e_i'\}$
\end{algorithmic}
\end{algorithm}

\section{Dataset Preparation}
\label{apx:datasets}

\subsection{Amorphous Silica (a-\ch{SiO2}) Dataset}
The a-\ch{SiO2} dataset contains $6,000$ samples that share the composition of pure silica, \ch{SiO2}.
To maximize the variation of properties between the samples generated with the same simulation workflow, relatively small unit cells are chosen with the number of atoms uniformly selected in the range of 80 to 250.
Atoms are initially placed in a unit cell with a volume $V = 4 \sum_i \frac{4}{3} \pi r_i^3$, with $r_i$ being the covalent radius of the $i$-th atom, avoiding unphysical overlap between neighboring atoms.
A local structure relaxation is performed on the initial configuration followed by an MD simulation in the NPT ensemble at 3500\,K for 2000\,ps.
To limit the effects of relaxation, which we observe for our other datasets, we use an instantaneous quenching procedure by performing a local structure optimization and a subsequent equilibration at 300\,K for 10\,ps.
All simulations are performed using LAMMPS~\citep{lammps} software and the Tersoff potential parameterized by \citep{munetoh2007interatomic}.

\subsection{Multi-Element Glass (MEG) Dataset}
The MEG dataset contains 9,027 multi-component glass samples with 11 different elements (Si, P, Al, Li, Be, K, Ca, Ti, Ba, Zn, O).
Compositions are generated from varying ratios of glass formers (\ch{SiO2}, \ch{P2O5}) and modifiers (\ch{Al2O3}, \ch{Li2O}, \ch{BeO}, \ch{K2O}, \ch{CaO}, \ch{TiO2}, \ch{BaO}, \ch{ZnO}), with up to four modifiers at 40\% total concentration relative to the glass formers.
Initial structures contain approximately 800 atoms placed randomly in a simulation cell with volume $V = 3 \sum_i \frac{4}{3} \pi r_i^3$, where $r_i$ is the covalent radius of atom $i$.
Samples are prepared using a melt-quench procedure: melting at $\frac{3}{4} T_\mathrm{evap}$ for 400\,ps, quenching to 300\,K at 5\,K/ps, and equilibrating for 300\,ps.
All simulations use LAMMPS~\citep{lammps} with the Bertani--Menziani--Pedone (BMP)-shrm potential~\citep{bertani2021improved}.

\subsection{Charge Balance of the Datasets}
\label{apx:dataset-charge}
Under the formal charge values in Appendix~\ref{apx:charge-assignment}, all 6,000 a-\ch{SiO2} samples and all 9,027 MEG samples have total charge $Q = 0$.
This follows from the construction of the datasets: the a-\ch{SiO2} samples have the fixed \ch{SiO2} stoichiometry, and the MEG compositions are assembled from the neutral oxide units listed above.

\section{Charge Balance}
\label{apx:charge}

\subsection{Charge Value Assignment}
\label{apx:charge-assignment}
The formal charge values $c_j$ in Definition~\ref{def:charge} are the integer oxidation states from the Bertani--Menziani--Pedone (BMP) potential~\citep{bertani2021improved}.
For the a-\ch{SiO2} dataset, the two element types have formal charges $c_{\mathrm{Si}} = +4$ and $c_{\mathrm{O}} = -2$.
For the MEG dataset, the 11 element types have the following formal charges: Si ($+4$), O ($-2$), P ($+5$), Al ($+3$), Li ($+1$), Be ($+2$), K ($+1$), Ca ($+2$), Ti ($+4$), Ba ($+2$), and Zn ($+2$).
Ghost atoms are assigned $c = 0$.

\subsection{Charge Balance Metrics}
\label{apx:charge-metrics}
Given $N$ generated samples with charge values $Q(\mE_i)$ for $i = 1, \ldots, N$, we report the following metrics.

\paragraph{Probability of charge balance.}
The percentage of generated samples satisfying $Q(\mE) = 0$.

\paragraph{Mean absolute charge value.}
\begin{equation}
|\bar{Q}| = \left|\frac{1}{N}\sum_{i=1}^{N} Q(\mE_i)\right|
\end{equation}

\paragraph{Standard deviation of charge value.}
\begin{equation}
\sigma_Q = \sqrt{\frac{1}{N}\sum_{i=1}^{N} (Q(\mE_i) - \bar{Q})^2}
\end{equation}

\subsection{Charge Balance and Stoichiometry}
\label{apx:stoichiometry}
In the training samples, zero total charge follows from how the compositions were constructed (Appendix~\ref{apx:dataset-charge}).
During generation, however, every element assignment is predicted, so unconstrained generation can produce element counts with nonzero total charge, and \archname{} enforces $Q = 0$ on each generated sample.
Let $N_j$ denote the number of atoms of element $j$ in a sample after removing ghost atoms.

For the a-\ch{SiO2} dataset, a sample contains only Si and O, and
\begin{equation}
Q = 4N_\mathrm{Si} - 2N_\mathrm{O} = 0 \quad\Longleftrightarrow\quad N_\mathrm{O} = 2N_\mathrm{Si}
\end{equation}
so formal charge balance is equivalent to the exact \ch{SiO2} stoichiometry.

For the MEG dataset, there is no single target formula, since the compositions vary over mixtures of the ten constituent oxides.
For any generated cation counts, these oxide units require
\begin{equation}
N_\mathrm{O}^\text{oxide} = 2N_\mathrm{Si} + \tfrac{5}{2}N_\mathrm{P} + \tfrac{3}{2}N_\mathrm{Al} + \tfrac{1}{2}N_\mathrm{Li} + N_\mathrm{Be} + \tfrac{1}{2}N_\mathrm{K} + N_\mathrm{Ca} + 2N_\mathrm{Ti} + N_\mathrm{Ba} + N_\mathrm{Zn}
\end{equation}
oxygen atoms, and under the formal charge values in Appendix~\ref{apx:charge-assignment}, $Q = 2(N_\mathrm{O}^\text{oxide} - N_\mathrm{O})$.
Formal charge balance is thus equivalent to a sample containing exactly the oxygen needed to form neutral oxide units with its cations.
The constraint leaves the cation ratios free, so the property-conditioned generation can select cation ratios beyond the compositions of the training samples.

Table~\ref{tab:stoichiometry} reports two rates over the samples generated by \archname{}: the stoichiometric validity rate, i.e., the percentage of samples satisfying the above criterion, and the training composition match rate, i.e., the percentage of samples whose simplest integer formula, obtained by dividing all element counts by their greatest common divisor, is identical to that of at least one training sample.
On the MEG dataset, the low match rates show that \archname{} usually generates new cation ratios instead of repeating the compositions of training samples.
This is expected for property-conditioned generation, especially for the target $C_\text{Li} = 15\%$, which lies outside the property distribution of the training samples (Figure~\ref{fig:dist-meg}).

\begin{table}[ht]
  \caption{Stoichiometric validity and training composition match rates of samples generated by \archname{}.}
  \label{tab:stoichiometry}
  \centering
  \begin{tabular}{l c c c}
    \toprule
    Config & Stoichiometric criterion & \shortstack{Stoichiometric\\validity rate} & \shortstack{Training composition\\match rate} \\
    \midrule
    a-\ch{SiO2}~$\mid$~$G$ & $N_\mathrm{O} = 2N_\mathrm{Si}$ & 100\% & 100\% \\
    a-\ch{SiO2}~$\mid$~RSD & $N_\mathrm{O} = 2N_\mathrm{Si}$ & 100\% & 100\% \\
    MEG~$\mid$~$E$+$C_\text{Li}$ & $N_\mathrm{O} = N_\mathrm{O}^\text{oxide}$ & 100\% & 0.00\% \\
    MEG~$\mid$~$G$ & $N_\mathrm{O} = N_\mathrm{O}^\text{oxide}$ & 100\% & 0.15\% \\
    \bottomrule
  \end{tabular}
\end{table}

\section{Inverse Design Evaluation}
\label{apx:inverse-design-eval}

\subsection{Property Calculation}
\label{apx:property-calculation}
We evaluate inverse design performance using material properties relevant to the dataset. For a-\ch{SiO2} samples, we focus on shear modulus and ring size distribution. For MEG samples, we focus on Young's modulus, shear modulus, and lithium molar concentration.

\paragraph{a-\ch{SiO2} dataset properties.}
For a-\ch{SiO2} samples, properties are computed directly from atomic structures. Shear modulus is calculated using finite differences of the stress tensor: structures are relaxed with the Tersoff potential (force tolerance 0.05 eV/\AA), then subjected to small strains ($\delta = 0.02$) to compute elastic constants $C_{44}$, $C_{55}$, and $C_{66}$ from stress responses. Shear modulus is their average: $G = \frac{1}{3}(C_{44} + C_{55} + C_{66})$. Ring size distribution is computed by identifying closed rings in the Si-O network using a depth-first search algorithm. Atoms are considered bonded if their distance is below 1.3 times the sum of their covalent radii (Si: 1.11 \AA, O: 0.66 \AA). The algorithm searches for the shortest path between bonded atom pairs while excluding the direct bond, ensuring proper ring closure with periodic boundary conditions. Ring sizes are reported as the number of Si atoms per ring, averaged across all identified rings.

\paragraph{MEG Dataset properties.}
For MEG samples, elastic properties are computed using the BMP-shrm potential~\cite{bertani2021improved}. Structures are first relaxed, then elastic constants $C_{ij}$ are calculated from finite differences of the stress tensor using Voigt notation. Young's modulus is computed as $E = \frac{(C_{11} - C_{12})(C_{11} + 2C_{12})}{C_{11} + C_{12}}$ and shear modulus as $G = C_{44}$. Lithium molar concentration $C_\text{Li}$ is computed directly from the elemental composition as the ratio of lithium atoms to total atoms.

\paragraph{Implementation details.}
Ghost atoms are excluded from all calculations. Structural relaxations use quasi-Newton optimization with variable cell shapes, maximum 1500 steps. Ring finding incorporates periodic boundary conditions for rings crossing cell boundaries.

\subsection{Inverse Design Metrics}
\label{apx:inverse-design-metrics}

For inverse design evaluation, we compare target properties $y_{\text{target}}$ with computed or predicted properties $y_{\text{generated}}$ of generated samples using three standard regression metrics:

\paragraph{Mean Absolute Error (MAE).}
\begin{equation}
\text{MAE} = \frac{1}{n} \sum_{i=1}^{n} |y_{\text{generated},i} - y_{\text{target},i}|
\end{equation}

\paragraph{Root Mean Square Error (RMSE).}
\begin{equation}
\text{RMSE} = \sqrt{\frac{1}{n} \sum_{i=1}^{n} (y_{\text{generated},i} - y_{\text{target},i})^2}
\end{equation}

\paragraph{Mean Absolute Percentage Error (MAPE).}
\begin{equation}
\text{MAPE} = \frac{100\%}{n} \sum_{i=1}^{n} \left|\frac{y_{\text{generated},i} - y_{\text{target},i}}{y_{\text{target},i}}\right|
\end{equation}

These metrics are computed separately for each property (shear modulus, RSD, Young's modulus, Li ratio) across all generated samples. MAE provides interpretable error magnitude in original units, RMSE penalizes large deviations more heavily, and MAPE enables comparison across properties with different scales and units.

\subsection{Training Data and Target Property Distributions}
\label{apx:property-distributions}
\begin{figure}[ht]
\centering
\begin{subfigure}[b]{\textwidth}
  \centering
  \begin{minipage}{0.32\textwidth}\centering\resizebox{\textwidth}{!}{%
\begin{tikzpicture}
\begin{axis}[
  width=6cm, height=5.0cm,
  xlabel={$G$ (GPa)},
  ylabel={Density ($\times 10^{-2}$)},
  xmin=8, xmax=52,
  xtick={10,20,30,40,50},
  ymin=0, ymax=18,
  y filter/.expression={y*100},
  yticklabel style={text width=1.2em, align=right},
  grid=major,
  grid style={gray!30},
  legend style={at={(0.97,0.97)}, anchor=north east, font=\scriptsize,
    fill=white, fill opacity=0.8, draw=none, text opacity=1},
]
\addplot[ybar interval, fill=plotblue, fill opacity=0.4, draw=plotblue!70]
  coordinates {
    (12.81,0.0) (13.59,0.0) (14.37,0.0) (15.15,0.0) (15.93,0.0)
    (16.72,0.0004) (17.5,0.0004) (18.28,0.0011) (19.06,0.0021) (19.84,0.0049)
    (20.62,0.0085) (21.41,0.0111) (22.19,0.0239) (22.97,0.0428) (23.75,0.0605)
    (24.53,0.1006) (25.32,0.1151) (26.1,0.1424) (26.88,0.1588) (27.66,0.1437)
    (28.44,0.1305) (29.23,0.103) (30.01,0.0727) (30.79,0.0516) (31.57,0.0362)
    (32.35,0.023) (33.14,0.0143) (33.92,0.0121) (34.7,0.0064) (35.48,0.006)
    (36.26,0.0028) (37.04,0.0017) (37.83,0.0006) (38.61,0.0006) (39.39,0.0002)
    (40.17,0.0) (40.95,0.0002) (41.74,0.0) (42.52,0.0006) (43.3,0)
  };
\addlegendentry{Training}
\addplot[ybar interval, fill=plotred, fill opacity=0.3, draw=plotred!70]
  coordinates {
    (12.81,0.0) (13.59,0.0093) (14.37,0.0093) (15.15,0.0232) (15.93,0.0364)
    (16.72,0.0325) (17.5,0.051) (18.28,0.0556) (19.06,0.0897) (19.84,0.0927)
    (20.62,0.0907) (21.41,0.0742) (22.19,0.0649) (22.97,0.0603) (23.75,0.0556)
    (24.53,0.0556) (25.32,0.0603) (26.1,0.0418) (26.88,0.0371) (27.66,0.0371)
    (28.44,0.0278) (29.23,0.0278) (30.01,0.0278) (30.79,0.0185) (31.57,0.0185)
    (32.35,0.0232) (33.14,0.0093) (33.92,0.0232) (34.7,0.0232) (35.48,0.0232)
    (36.26,0.0093) (37.04,0.014) (37.83,0.0093) (38.61,0.01) (39.39,0.0093)
    (40.17,0.0093) (40.95,0.01) (41.74,0.003) (42.52,0.0047) (43.3,0)
  };
\addlegendentry{\archname{}}
\draw[gray, dashed, thick] (axis cs:10,0) -- (axis cs:10,18);
\draw[gray, dashed, thick] (axis cs:50,0) -- (axis cs:50,18);
\end{axis}
\end{tikzpicture}%
}\end{minipage}
  \hspace{1em}
  \begin{minipage}{0.32\textwidth}\centering\resizebox{\textwidth}{!}{%
\begin{tikzpicture}
\begin{axis}[
  width=6cm, height=5.0cm,
  xlabel={RSD (atoms)},
  ylabel={Density ($\times 10^{-2}$)},
  xmin=3.8, xmax=6.2,
  xtick={4.0,4.5,5.0,5.5,6.0},
  ymin=0, ymax=260,
  y filter/.expression={y*100},
  yticklabel style={text width=1.2em, align=right},
  grid=major,
  grid style={gray!30},
  legend style={at={(0.97,0.97)}, anchor=north east, font=\scriptsize,
    fill=white, fill opacity=0.8, draw=none, text opacity=1},
]
\addplot[ybar interval, fill=plotblue, fill opacity=0.4, draw=plotblue!70]
  coordinates {
    (3.995,0.0) (4.037,0.004) (4.079,0.0) (4.121,0.004) (4.163,0.0079)
    (4.205,0.0158) (4.247,0.0079) (4.289,0.0396) (4.331,0.0554) (4.373,0.0554)
    (4.415,0.1069) (4.458,0.1544) (4.5,0.3246) (4.542,0.4434) (4.584,0.574)
    (4.626,0.9066) (4.668,0.962) (4.71,1.3777) (4.752,1.5994) (4.794,1.9042)
    (4.836,2.0745) (4.879,2.2447) (4.921,2.1062) (4.963,2.3041) (5.005,1.5559)
    (5.047,1.5677) (5.089,1.1441) (5.131,0.8551) (5.173,0.5345) (5.215,0.3286)
    (5.257,0.1742) (5.3,0.1742) (5.342,0.0831) (5.384,0.0356) (5.426,0.0158)
    (5.468,0.0) (5.51,0.0079) (5.552,0.0) (5.594,0.004) (5.636,0)
  };
\addplot[ybar interval, fill=plotred, fill opacity=0.3, draw=plotred!70]
  coordinates {
    (3.995,0.305) (4.037,0.2288) (4.079,0.2288) (4.121,0.4046) (4.163,0.305)
    (4.205,0.3813) (4.247,0.3813) (4.289,0.5307) (4.331,0.3813) (4.373,0.5043)
    (4.415,0.3813) (4.458,0.8389) (4.5,0.9915) (4.542,0.7393) (4.584,1.1439)
    (4.626,0.9152) (4.668,0.9152) (4.71,0.9152) (4.752,1.0677) (4.794,1.2203)
    (4.836,0.7626) (4.879,1.4491) (4.921,1.3727) (4.963,1.6779) (5.005,1.9829)
    (5.047,1.1439) (5.089,0.9152) (5.131,0.6864) (5.173,0.3813) (5.215,0.2288)
    (5.257,0.2288) (5.3,0.0762) (5.342,0.0411) (5.384,0.0226) (5.426,0.0113)
    (5.468,0.0164) (5.51,0.0) (5.552,0.0) (5.594,0.0) (5.636,0)
  };
\draw[gray, dashed, thick] (axis cs:4,0) -- (axis cs:4,260);
\draw[gray, dashed, thick] (axis cs:6,0) -- (axis cs:6,260);
\end{axis}
\end{tikzpicture}%
}\end{minipage}
  \caption{a-\ch{SiO2}~$\mid$~$G$ and a-\ch{SiO2}~$\mid$~RSD configurations}
  \label{fig:dist-sio2}
\end{subfigure}
\\[1em]
\begin{subfigure}[b]{\textwidth}
  \centering
  \begin{minipage}{0.32\textwidth}\centering\resizebox{\textwidth}{!}{%
\begin{tikzpicture}
\begin{axis}[
  width=6cm, height=5.0cm,
  xlabel={$E$ (GPa)},
  ylabel={Density ($\times 10^{-2}$)},
  xmin=0, xmax=170,
  xtick={0,50,100,150},
  ymin=0, ymax=3.2,
  y filter/.expression={y*100},
  yticklabel style={text width=1.2em, align=right},
  grid=major,
  grid style={gray!30},
  legend style={at={(0.97,0.97)}, anchor=north east, font=\scriptsize,
    fill=white, fill opacity=0.8, draw=none, text opacity=1},
]
\addplot[ybar interval, fill=plotblue, fill opacity=0.4, draw=plotblue!70]
  coordinates {
    (16.81,0.0) (20.32,0.0) (23.84,0.0) (27.35,0.0) (30.87,0.0001)
    (34.38,0.0001) (37.9,0.0008) (41.42,0.0015) (44.93,0.0037) (48.45,0.0072)
    (51.96,0.0109) (55.48,0.0161) (58.99,0.0207) (62.51,0.0233) (66.02,0.0271)
    (69.54,0.0259) (73.05,0.0257) (76.57,0.0238) (80.09,0.019) (83.6,0.0168)
    (87.12,0.013) (90.63,0.0114) (94.15,0.0074) (97.66,0.0076) (101.2,0.006)
    (104.7,0.0043) (108.2,0.0034) (111.7,0.0028) (115.2,0.0025) (118.8,0.0012)
    (122.3,0.0007) (125.8,0.0006) (129.3,0.0006) (132.8,0.0002) (136.3,0.0)
    (139.8,0.0) (143.4,0.0001) (146.9,0.0) (150.4,0.0) (153.9,0)
  };
\addplot[ybar interval, fill=plotred, fill opacity=0.3, draw=plotred!70]
  coordinates {
    (16.81,0.0009) (20.32,0.0017) (23.84,0.0066) (27.35,0.0131) (30.87,0.0125)
    (34.38,0.0105) (37.9,0.0106) (41.42,0.0099) (44.93,0.0105) (48.45,0.0067)
    (51.96,0.0083) (55.48,0.0096) (58.99,0.006) (62.51,0.0091) (66.02,0.0083)
    (69.54,0.0085) (73.05,0.0083) (76.57,0.0066) (80.09,0.0066) (83.6,0.008)
    (87.12,0.0085) (90.63,0.0088) (94.15,0.0083) (97.66,0.0134) (101.2,0.0146)
    (104.7,0.0131) (108.2,0.013) (111.7,0.0108) (115.2,0.0131) (118.8,0.0108)
    (122.3,0.0031) (125.8,0.0017) (129.3,0.001) (132.8,0.0013) (136.3,0.0009)
    (139.8,0.0) (143.4,0.0) (146.9,0.0) (150.4,0.0) (153.9,0)
  };
\draw[gray, dashed, thick] (axis cs:20,0) -- (axis cs:20,3.2);
\draw[gray, dashed, thick] (axis cs:160,0) -- (axis cs:160,3.2);
\end{axis}
\end{tikzpicture}%
}\end{minipage}
  \hfill
  \begin{minipage}{0.32\textwidth}\centering\resizebox{\textwidth}{!}{%
\begin{tikzpicture}
\begin{axis}[
  width=6cm, height=5.0cm,
  xlabel={$C_\text{Li}$},
  ylabel={Density},
  xmin=0, xmax=0.2,
  xtick={0,0.05,0.10,0.15,0.20},
  xticklabel style={/pgf/number format/fixed},
  ymin=0, ymax=140,
  yticklabel style={text width=1.2em, align=right},
  grid=major,
  grid style={gray!30},
]
\addplot[ybar interval, fill=plotblue, fill opacity=0.4, draw=plotblue!70]
  coordinates {
    (0.0,126.5455) (0.005011,0.0) (0.01002,0.0) (0.01503,0.0) (0.02005,0.0)
    (0.02506,0.0) (0.03007,5.6817) (0.03508,32.4544) (0.04009,12.4467) (0.0451,0.3758)
    (0.05011,0.0) (0.05512,0.0) (0.06014,0.0442) (0.06515,0.7959) (0.07016,3.3383)
    (0.07517,6.6324) (0.08018,4.3111) (0.08519,1.5254) (0.0902,0.4863) (0.09521,0.0221)
    (0.1002,0.0442) (0.1052,0.2432) (0.1102,0.7738) (0.1153,0.9949) (0.1203,0.9285)
    (0.1253,0.7296) (0.1303,0.2874) (0.1353,0.2653) (0.1403,0.0221) (0.1453,0.0221)
    (0.1503,0.0663) (0.1553,0.1105) (0.1604,0.0663) (0.1654,0.1547) (0.1704,0.0663)
    (0.1754,0.0442) (0.1804,0.0442) (0.1854,0.0221) (0.1904,0.0221) (0.1954,0)
  };
\addplot[ybar interval, fill=plotred, fill opacity=0.3, draw=plotred!70]
  coordinates {
    (0.0061,0.0) (0.0111,0.0) (0.0161,0.0) (0.0211,0.0) (0.0261,0.0)
    (0.0312,0.0) (0.0362,0.0) (0.0412,0.0) (0.0462,0.0) (0.0512,0.0)
    (0.0562,0.0) (0.0612,0.0) (0.0662,0.0) (0.0712,0.0) (0.0763,0.0)
    (0.0813,0.0) (0.0863,0.0) (0.0913,0.0) (0.0963,0.0) (0.1013,0.0)
    (0.1063,0.3045) (0.1113,1.2179) (0.1163,1.8269) (0.1214,8.5252) (0.1264,9.1341)
    (0.1314,20.3995) (0.1364,20.3995) (0.1414,26.4888) (0.1464,24.0531) (0.1514,24.662)
    (0.1564,18.2681) (0.1614,11.5698) (0.1665,11.2654) (0.1715,8.2207) (0.1765,7.3072)
    (0.1815,3.6536) (0.1865,0.9134) (0.1915,0.9134) (0.1965,0.609) (0.2015,0)
  };
\draw[gray, dashed, thick] (axis cs:0.15,0) -- (axis cs:0.15,140);
\end{axis}
\end{tikzpicture}%
}\end{minipage}
  \hfill
  \begin{minipage}{0.32\textwidth}\centering\resizebox{\textwidth}{!}{%
\begin{tikzpicture}
\begin{axis}[
  width=6cm, height=5.0cm,
  xlabel={$G$ (GPa)},
  ylabel={Density ($\times 10^{-2}$)},
  xmin=0, xmax=75,
  xtick={0,20,40,60},
  ymin=0, ymax=8,
  y filter/.expression={y*100},
  yticklabel style={text width=1.2em, align=right},
  grid=major,
  grid style={gray!30},
  legend style={at={(0.97,0.97)}, anchor=north east, font=\scriptsize,
    fill=white, fill opacity=0.8, draw=none, text opacity=1},
]
\addplot[ybar interval, fill=plotblue, fill opacity=0.4, draw=plotblue!70]
  coordinates {
    (6.282,0.0) (7.625,0.0) (8.969,0.0) (10.31,0.0) (11.66,0.0)
    (13.0,0.0) (14.34,0.0005) (15.69,0.0018) (17.03,0.004) (18.38,0.0079)
    (19.72,0.0171) (21.06,0.0223) (22.41,0.0326) (23.75,0.045) (25.09,0.0555)
    (26.44,0.0623) (27.78,0.0658) (29.13,0.0682) (30.47,0.063) (31.81,0.0552)
    (33.16,0.048) (34.5,0.0393) (35.84,0.0357) (37.19,0.0254) (38.53,0.0227)
    (39.88,0.0171) (41.22,0.0151) (42.56,0.0115) (43.91,0.0092) (45.25,0.0058)
    (46.6,0.004) (47.94,0.0031) (49.28,0.0031) (50.63,0.0012) (51.97,0.0008)
    (53.31,0.0006) (54.66,0.0002) (56.0,0.0001) (57.35,0.0002) (58.69,0)
  };
\addplot[ybar interval, fill=plotred, fill opacity=0.3, draw=plotred!70]
  coordinates {
    (6.282,0.0038) (7.625,0.0038) (8.969,0.0121) (10.31,0.022) (11.66,0.0288)
    (13.0,0.0341) (14.34,0.028) (15.69,0.0265) (17.03,0.0257) (18.38,0.0151)
    (19.72,0.0182) (21.06,0.0197) (22.41,0.022) (23.75,0.0167) (25.09,0.0212)
    (26.44,0.0197) (27.78,0.0273) (29.13,0.0182) (30.47,0.0114) (31.81,0.0212)
    (33.16,0.0227) (34.5,0.0189) (35.84,0.0227) (37.19,0.0204) (38.53,0.025)
    (39.88,0.0333) (41.22,0.0295) (42.56,0.0333) (43.91,0.025) (45.25,0.0288)
    (46.6,0.0326) (47.94,0.028) (49.28,0.0189) (50.63,0.0068) (51.97,0.0023)
    (53.31,0.0008) (54.66,0.0) (56.0,0.0) (57.35,0.0) (58.69,0)
  };
\draw[gray, dashed, thick] (axis cs:10,0) -- (axis cs:10,8);
\draw[gray, dashed, thick] (axis cs:70,0) -- (axis cs:70,8);
\end{axis}
\end{tikzpicture}%
}\end{minipage}
  \caption{MEG~$\mid$~$E$+$C_\text{Li}$ and MEG~$\mid$~$G$ configurations}
  \label{fig:dist-meg}
\end{subfigure}
\caption{Property distributions of training data and samples generated by \archname{} for each inverse design configuration. Dashed lines indicate the target generation range (or target value for $C_\text{Li}$).}
\label{fig:property-dist}
\end{figure}
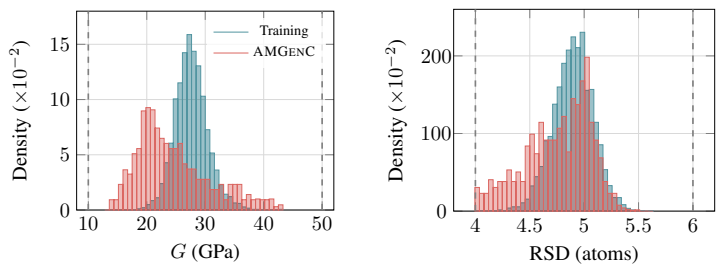
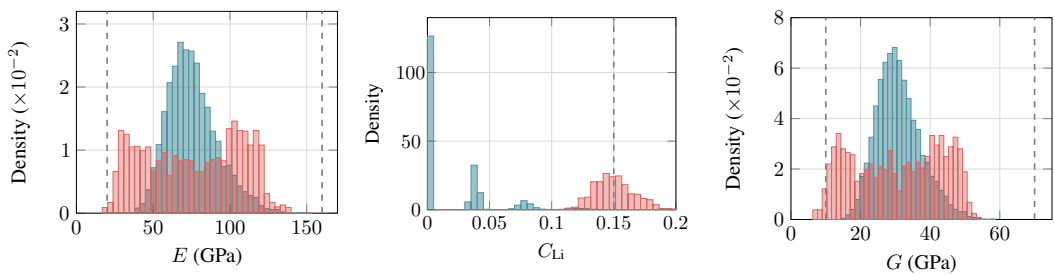

Figure~\ref{fig:property-dist} compares the property distributions of training data and samples generated by \archname{} for each inverse design configuration.
The \archname{} distributions extend beyond the training data range, demonstrating that the model is capable of extrapolating to property values not seen during training.

On the other hand, it is shown that the generated samples extrapolate more readily toward lower property values than toward higher ones.
This reflects physical constraints of the material systems: on the MEG dataset, network modifiers such as Li reduce network connectivity and thus stiffness~\citep{rouxel2007elastic}, making very high elastic moduli difficult to achieve at the targeted Li content ($C_\text{Li} = 0.15$).
Similarly, for a-\ch{SiO2}, the achievable range of shear modulus and ring size distribution is bounded by the structural constraints of the silica network.

\section{Structural Features of Generated Samples}
\label{apx:structural-features}

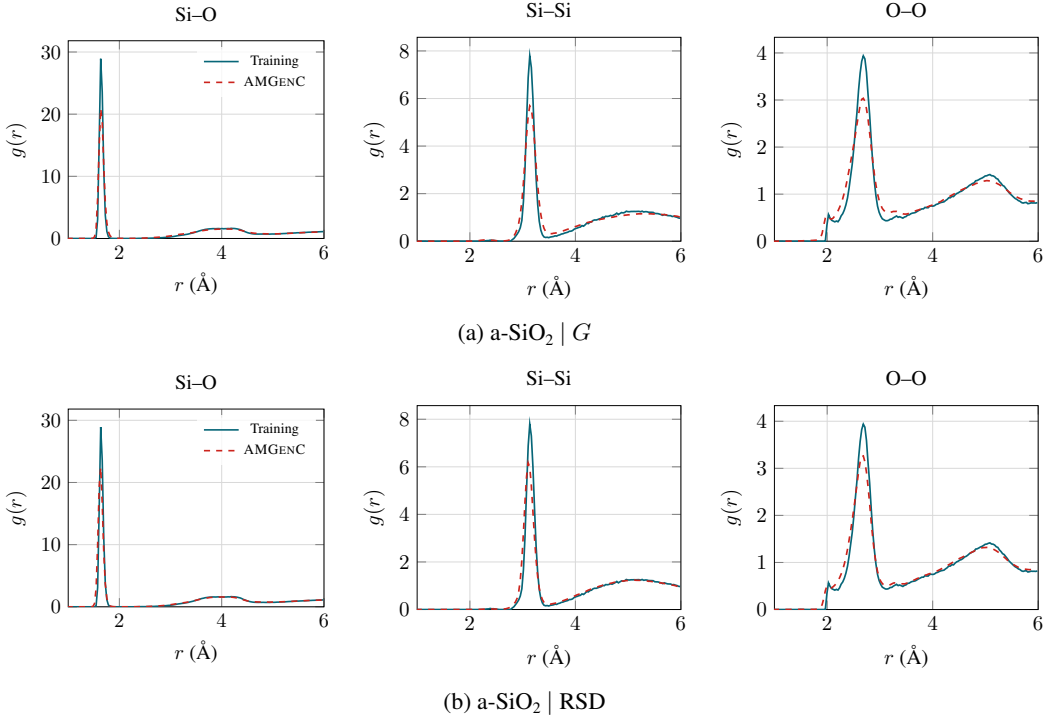
\begin{figure}[ht]
\centering
\begin{subfigure}[b]{\textwidth}
  \centering
  \begin{minipage}{0.32\textwidth}\centering\resizebox{\textwidth}{!}{%
\begin{tikzpicture}
\begin{axis}[
  width=6cm, height=5.0cm,
  xlabel={$r$ (\AA)},
  ylabel={$g(r)$},
  xmin=1.0, xmax=6.0,
  ymin=0,
  title={Si--O},
  grid=major,
  grid style={gray!30},
  legend style={at={(0.97,0.97)}, anchor=north east, font=\scriptsize,
    fill=white, fill opacity=0.8, draw=none, text opacity=1},
]
\addplot[plotblue, thick, no markers]
  table[x=r, y=train] {comp/struct_feat/sio2_rdf_SiO_condG.dat};
\addlegendentry{Training}
\addplot[plotred, thick, dashed, no markers]
  table[x=r, y=amcharge] {comp/struct_feat/sio2_rdf_SiO_condG.dat};
\addlegendentry{\archname{}}
\end{axis}
\end{tikzpicture}%
}\end{minipage}
  \hfill
  \begin{minipage}{0.32\textwidth}\centering\resizebox{\textwidth}{!}{%
\begin{tikzpicture}
\begin{axis}[
  width=6cm, height=5.0cm,
  xlabel={$r$ (\AA)},
  ylabel={$g(r)$},
  xmin=1.0, xmax=6.0,
  ymin=0,
  title={Si--Si},
  grid=major,
  grid style={gray!30},
]
\addplot[plotblue, thick, no markers]
  table[x=r, y=train] {comp/struct_feat/sio2_rdf_SiSi_condG.dat};
\addplot[plotred, thick, dashed, no markers]
  table[x=r, y=amcharge] {comp/struct_feat/sio2_rdf_SiSi_condG.dat};
\end{axis}
\end{tikzpicture}%
}\end{minipage}
  \hfill
  \begin{minipage}{0.32\textwidth}\centering\resizebox{\textwidth}{!}{%
\begin{tikzpicture}
\begin{axis}[
  width=6cm, height=5.0cm,
  xlabel={$r$ (\AA)},
  ylabel={$g(r)$},
  xmin=1.0, xmax=6.0,
  ymin=0,
  title={O--O},
  grid=major,
  grid style={gray!30},
]
\addplot[plotblue, thick, no markers]
  table[x=r, y=train] {comp/struct_feat/sio2_rdf_OO_condG.dat};
\addplot[plotred, thick, dashed, no markers]
  table[x=r, y=amcharge] {comp/struct_feat/sio2_rdf_OO_condG.dat};
\end{axis}
\end{tikzpicture}%
}\end{minipage}
  \caption{a-\ch{SiO2}~$\mid$~$G$}
\end{subfigure}
\\[0.5em]
\begin{subfigure}[b]{\textwidth}
  \centering
  \begin{minipage}{0.32\textwidth}\centering\resizebox{\textwidth}{!}{%
\begin{tikzpicture}
\begin{axis}[
  width=6cm, height=5.0cm,
  xlabel={$r$ (\AA)},
  ylabel={$g(r)$},
  xmin=1.0, xmax=6.0,
  ymin=0,
  title={Si--O},
  grid=major,
  grid style={gray!30},
  legend style={at={(0.97,0.97)}, anchor=north east, font=\scriptsize,
    fill=white, fill opacity=0.8, draw=none, text opacity=1},
]
\addplot[plotblue, thick, no markers]
  table[x=r, y=train] {comp/struct_feat/sio2_rdf_SiO_condRSD.dat};
\addlegendentry{Training}
\addplot[plotred, thick, dashed, no markers]
  table[x=r, y=amcharge] {comp/struct_feat/sio2_rdf_SiO_condRSD.dat};
\addlegendentry{\archname{}}
\end{axis}
\end{tikzpicture}%
}\end{minipage}
  \hfill
  \begin{minipage}{0.32\textwidth}\centering\resizebox{\textwidth}{!}{%
\begin{tikzpicture}
\begin{axis}[
  width=6cm, height=5.0cm,
  xlabel={$r$ (\AA)},
  ylabel={$g(r)$},
  xmin=1.0, xmax=6.0,
  ymin=0,
  title={Si--Si},
  grid=major,
  grid style={gray!30},
]
\addplot[plotblue, thick, no markers]
  table[x=r, y=train] {comp/struct_feat/sio2_rdf_SiSi_condRSD.dat};
\addplot[plotred, thick, dashed, no markers]
  table[x=r, y=amcharge] {comp/struct_feat/sio2_rdf_SiSi_condRSD.dat};
\end{axis}
\end{tikzpicture}%
}\end{minipage}
  \hfill
  \begin{minipage}{0.32\textwidth}\centering\resizebox{\textwidth}{!}{%
\begin{tikzpicture}
\begin{axis}[
  width=6cm, height=5.0cm,
  xlabel={$r$ (\AA)},
  ylabel={$g(r)$},
  xmin=1.0, xmax=6.0,
  ymin=0,
  title={O--O},
  grid=major,
  grid style={gray!30},
]
\addplot[plotblue, thick, no markers]
  table[x=r, y=train] {comp/struct_feat/sio2_rdf_OO_condRSD.dat};
\addplot[plotred, thick, dashed, no markers]
  table[x=r, y=amcharge] {comp/struct_feat/sio2_rdf_OO_condRSD.dat};
\end{axis}
\end{tikzpicture}%
}\end{minipage}
  \caption{a-\ch{SiO2}~$\mid$~RSD}
\end{subfigure}
\caption{Partial radial distribution functions $g(r)$ for a-\ch{SiO2} samples.
Solid: training samples.
Dashed: samples generated by \archname{} under the (a)~a-\ch{SiO2}~$\mid$~$G$ configuration and (b)~a-\ch{SiO2}~$\mid$~RSD configuration.}
\label{fig:sio2-partial-rdf}
\end{figure}

Although structural accuracy is not the primary focus of this work, we compare key structural features of samples generated by \archname{} with those of the training samples.
Figures~\ref{fig:sio2-partial-rdf} and~\ref{fig:meg-partial-rdf} compare the partial radial distribution functions $g(r)$ of generated and training samples.
Overall, the generated samples reproduce the correct first-shell peak positions across all element pairs.
In some cases the peaks are lower and broader than those of the training data.
This can be attributed to two factors: the generation is conditioned on target properties that extend beyond the training distribution (Appendix~\ref{apx:property-distributions}), leading to compositional shifts between generated and training samples; and diffusion and flow matching models have inherent limitations in fully reproducing the well-relaxed local structures of amorphous materials~\citep{finkler2025inverse}.
Note that for the a-\ch{SiO2} dataset, the training curves may not appear perfectly smooth due to the relatively small simulation cells in the training data.
For the MEG dataset, some plots appear noisier than others due to the low abundance of certain network modifiers in the glass compositions.

\begin{figure}[!t]
\centering
\begin{tikzpicture}
\begin{axis}[
  width=12cm, height=8cm,
  xlabel={$r$ (\AA)},
  ylabel={$n(r)$},
  xmin=1.0, xmax=4.0,
  ymin=-0.2, ymax=8.2,
  grid=major,
  grid style={gray!30},
  legend style={at={(0.97,0.97)}, anchor=north east, font=\scriptsize,
    fill=white, fill opacity=0.8, draw=none, text opacity=1,
    legend columns=2},
]
\addplot[plotblue, thick, no markers]
  table[x=r, y=train] {comp/struct_feat/meg_cum_cn_Si.dat};
\addlegendentry{Si}
\addplot[plotblue, thick, dashed, no markers, forget plot]
  table[x=r, y=amcharge] {comp/struct_feat/meg_cum_cn_Si.dat};
\addplot[gruvorange, thick, no markers]
  table[x=r, y=train] {comp/struct_feat/meg_cum_cn_P.dat};
\addlegendentry{P}
\addplot[gruvorange, thick, dashed, no markers, forget plot]
  table[x=r, y=amcharge] {comp/struct_feat/meg_cum_cn_P.dat};
\addplot[gruvaqua, thick, no markers]
  table[x=r, y=train] {comp/struct_feat/meg_cum_cn_Al.dat};
\addlegendentry{Al}
\addplot[gruvaqua, thick, dashed, no markers, forget plot]
  table[x=r, y=amcharge] {comp/struct_feat/meg_cum_cn_Al.dat};
\addplot[plotred, thick, no markers]
  table[x=r, y=train] {comp/struct_feat/meg_cum_cn_Li.dat};
\addlegendentry{Li}
\addplot[plotred, thick, dashed, no markers, forget plot]
  table[x=r, y=amcharge] {comp/struct_feat/meg_cum_cn_Li.dat};
\addplot[gruvpurple, thick, no markers]
  table[x=r, y=train] {comp/struct_feat/meg_cum_cn_Ti.dat};
\addlegendentry{Ti}
\addplot[gruvpurple, thick, dashed, no markers, forget plot]
  table[x=r, y=amcharge] {comp/struct_feat/meg_cum_cn_Ti.dat};
\addplot[gruvyellow, thick, no markers]
  table[x=r, y=train] {comp/struct_feat/meg_cum_cn_Ca.dat};
\addlegendentry{Ca}
\addplot[gruvyellow, thick, dashed, no markers, forget plot]
  table[x=r, y=amcharge] {comp/struct_feat/meg_cum_cn_Ca.dat};
\addplot[gruvgreen, thick, no markers]
  table[x=r, y=train] {comp/struct_feat/meg_cum_cn_Ba.dat};
\addlegendentry{Ba}
\addplot[gruvgreen, thick, dashed, no markers, forget plot]
  table[x=r, y=amcharge] {comp/struct_feat/meg_cum_cn_Ba.dat};
\addplot[gruvbrightorange, thick, no markers]
  table[x=r, y=train] {comp/struct_feat/meg_cum_cn_Be.dat};
\addlegendentry{Be}
\addplot[gruvbrightorange, thick, dashed, no markers, forget plot]
  table[x=r, y=amcharge] {comp/struct_feat/meg_cum_cn_Be.dat};
\addplot[gruvbrightaqua, thick, no markers]
  table[x=r, y=train] {comp/struct_feat/meg_cum_cn_K.dat};
\addlegendentry{K}
\addplot[gruvbrightaqua, thick, dashed, no markers, forget plot]
  table[x=r, y=amcharge] {comp/struct_feat/meg_cum_cn_K.dat};
\addplot[gruvgray, thick, no markers]
  table[x=r, y=train] {comp/struct_feat/meg_cum_cn_Zn.dat};
\addlegendentry{Zn}
\addplot[gruvgray, thick, dashed, no markers, forget plot]
  table[x=r, y=amcharge] {comp/struct_feat/meg_cum_cn_Zn.dat};
\end{axis}
\end{tikzpicture}%
\caption{Cumulative coordination number $n(r)$ of oxygen neighbors around each
cation type in the MEG samples. Solid: training samples. Dashed: samples generated by \archname{} under the MEG~$\mid$~$E$+$C_\text{Li}$ configuration.}
\label{fig:meg-cum-cn}
\end{figure}
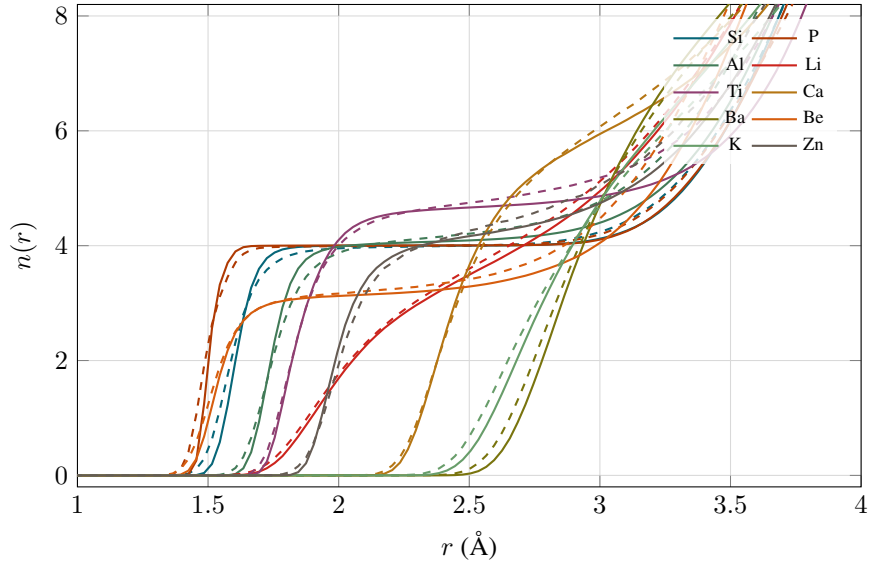

Figure~\ref{fig:meg-cum-cn} compares the cumulative coordination numbers $n(r)$ of oxygen neighbors around each cation type in the MEG samples.
The first-shell plateau values confirm that the generated structures reproduce the expected coordination numbers.
The agreement is strongest in the first coordination shell.
Beyond the first shell, minor deviations appear, consistent with the compositional differences between the conditioned generation and the training distribution.

\begin{figure}[!t]
\centering
\begin{minipage}{0.24\textwidth}\centering\resizebox{\textwidth}{!}{%
\begin{tikzpicture}
\begin{axis}[
  width=6cm, height=5.0cm,
  xlabel={$r$ (\AA)},
  ylabel={$g(r)$},
  xmin=1.0, xmax=6.0,
  ymin=0,
  title={Si--O},
  grid=major,
  grid style={gray!30},
  legend style={at={(0.97,0.97)}, anchor=north east, font=\scriptsize,
    fill=white, fill opacity=0.8, draw=none, text opacity=1},
]
\addplot[plotblue, thick, no markers]
  table[x=r, y=train] {comp/struct_feat/meg_rdf_SiO.dat};
\addlegendentry{Training}
\addplot[plotred, thick, dashed, no markers]
  table[x=r, y=amcharge] {comp/struct_feat/meg_rdf_SiO.dat};
\addlegendentry{\archname{}}
\end{axis}
\end{tikzpicture}%
}\end{minipage}
\hfill
\begin{minipage}{0.24\textwidth}\centering\resizebox{\textwidth}{!}{%
\begin{tikzpicture}
\begin{axis}[
  width=6cm, height=5.0cm,
  xlabel={$r$ (\AA)},
  ylabel={$g(r)$},
  xmin=1.0, xmax=6.0,
  ymin=0,
  title={P--O},
  grid=major,
  grid style={gray!30},
]
\addplot[plotblue, thick, no markers]
  table[x=r, y=train] {comp/struct_feat/meg_rdf_PO.dat};
\addplot[plotred, thick, dashed, no markers]
  table[x=r, y=amcharge] {comp/struct_feat/meg_rdf_PO.dat};
\end{axis}
\end{tikzpicture}%
}\end{minipage}
\hfill
\begin{minipage}{0.24\textwidth}\centering\resizebox{\textwidth}{!}{%
\begin{tikzpicture}
\begin{axis}[
  width=6cm, height=5.0cm,
  xlabel={$r$ (\AA)},
  ylabel={$g(r)$},
  xmin=1.0, xmax=6.0,
  ymin=0,
  title={Al--O},
  grid=major,
  grid style={gray!30},
]
\addplot[plotblue, thick, no markers]
  table[x=r, y=train] {comp/struct_feat/meg_rdf_AlO.dat};
\addplot[plotred, thick, dashed, no markers]
  table[x=r, y=amcharge] {comp/struct_feat/meg_rdf_AlO.dat};
\end{axis}
\end{tikzpicture}%
}\end{minipage}
\hfill
\begin{minipage}{0.24\textwidth}\centering\resizebox{\textwidth}{!}{%
\begin{tikzpicture}
\begin{axis}[
  width=6cm, height=5.0cm,
  xlabel={$r$ (\AA)},
  ylabel={$g(r)$},
  xmin=1.0, xmax=6.0,
  ymin=0,
  title={Li--O},
  grid=major,
  grid style={gray!30},
]
\addplot[plotblue, thick, no markers]
  table[x=r, y=train] {comp/struct_feat/meg_rdf_LiO.dat};
\addplot[plotred, thick, dashed, no markers]
  table[x=r, y=amcharge] {comp/struct_feat/meg_rdf_LiO.dat};
\end{axis}
\end{tikzpicture}%
}\end{minipage}
\\[0.5em]
\begin{minipage}{0.24\textwidth}\centering\resizebox{\textwidth}{!}{%
\begin{tikzpicture}
\begin{axis}[
  width=6cm, height=5.0cm,
  xlabel={$r$ (\AA)},
  ylabel={$g(r)$},
  xmin=1.0, xmax=6.0,
  ymin=0,
  title={Be--O},
  grid=major,
  grid style={gray!30},
]
\addplot[plotblue, thick, no markers]
  table[x=r, y=train] {comp/struct_feat/meg_rdf_BeO.dat};
\addplot[plotred, thick, dashed, no markers]
  table[x=r, y=amcharge] {comp/struct_feat/meg_rdf_BeO.dat};
\end{axis}
\end{tikzpicture}%
}\end{minipage}
\hfill
\begin{minipage}{0.24\textwidth}\centering\resizebox{\textwidth}{!}{%
\begin{tikzpicture}
\begin{axis}[
  width=6cm, height=5.0cm,
  xlabel={$r$ (\AA)},
  ylabel={$g(r)$},
  xmin=1.0, xmax=6.0,
  ymin=0,
  title={K--O},
  grid=major,
  grid style={gray!30},
]
\addplot[plotblue, thick, no markers]
  table[x=r, y=train] {comp/struct_feat/meg_rdf_KO.dat};
\addplot[plotred, thick, dashed, no markers]
  table[x=r, y=amcharge] {comp/struct_feat/meg_rdf_KO.dat};
\end{axis}
\end{tikzpicture}%
}\end{minipage}
\hfill
\begin{minipage}{0.24\textwidth}\centering\resizebox{\textwidth}{!}{%
\begin{tikzpicture}
\begin{axis}[
  width=6cm, height=5.0cm,
  xlabel={$r$ (\AA)},
  ylabel={$g(r)$},
  xmin=1.0, xmax=6.0,
  ymin=0,
  title={Ca--O},
  grid=major,
  grid style={gray!30},
]
\addplot[plotblue, thick, no markers]
  table[x=r, y=train] {comp/struct_feat/meg_rdf_CaO.dat};
\addplot[plotred, thick, dashed, no markers]
  table[x=r, y=amcharge] {comp/struct_feat/meg_rdf_CaO.dat};
\end{axis}
\end{tikzpicture}%
}\end{minipage}
\hfill
\begin{minipage}{0.24\textwidth}\centering\resizebox{\textwidth}{!}{%
\begin{tikzpicture}
\begin{axis}[
  width=6cm, height=5.0cm,
  xlabel={$r$ (\AA)},
  ylabel={$g(r)$},
  xmin=1.0, xmax=6.0,
  ymin=0,
  title={Ti--O},
  grid=major,
  grid style={gray!30},
]
\addplot[plotblue, thick, no markers]
  table[x=r, y=train] {comp/struct_feat/meg_rdf_TiO.dat};
\addplot[plotred, thick, dashed, no markers]
  table[x=r, y=amcharge] {comp/struct_feat/meg_rdf_TiO.dat};
\end{axis}
\end{tikzpicture}%
}\end{minipage}
\\[0.5em]
\begin{minipage}{0.24\textwidth}\centering\resizebox{\textwidth}{!}{%
\begin{tikzpicture}
\begin{axis}[
  width=6cm, height=5.0cm,
  xlabel={$r$ (\AA)},
  ylabel={$g(r)$},
  xmin=1.0, xmax=6.0,
  ymin=0,
  title={Ba--O},
  grid=major,
  grid style={gray!30},
]
\addplot[plotblue, thick, no markers]
  table[x=r, y=train] {comp/struct_feat/meg_rdf_BaO.dat};
\addplot[plotred, thick, dashed, no markers]
  table[x=r, y=amcharge] {comp/struct_feat/meg_rdf_BaO.dat};
\end{axis}
\end{tikzpicture}%
}\end{minipage}
\hspace{.5cm}
\begin{minipage}{0.24\textwidth}\centering\resizebox{\textwidth}{!}{%
\begin{tikzpicture}
\begin{axis}[
  width=6cm, height=5.0cm,
  xlabel={$r$ (\AA)},
  ylabel={$g(r)$},
  xmin=1.0, xmax=6.0,
  ymin=0,
  title={Zn--O},
  grid=major,
  grid style={gray!30},
]
\addplot[plotblue, thick, no markers]
  table[x=r, y=train] {comp/struct_feat/meg_rdf_ZnO.dat};
\addplot[plotred, thick, dashed, no markers]
  table[x=r, y=amcharge] {comp/struct_feat/meg_rdf_ZnO.dat};
\end{axis}
\end{tikzpicture}%
}\end{minipage}
\caption{Partial radial distribution functions $g(r)$ for element--oxygen pairs
in the MEG samples. Solid: training samples. Dashed: samples generated by \archname{} under the MEG~$\mid$~$E$+$C_\text{Li}$ configuration.}
\label{fig:meg-partial-rdf}
\end{figure}
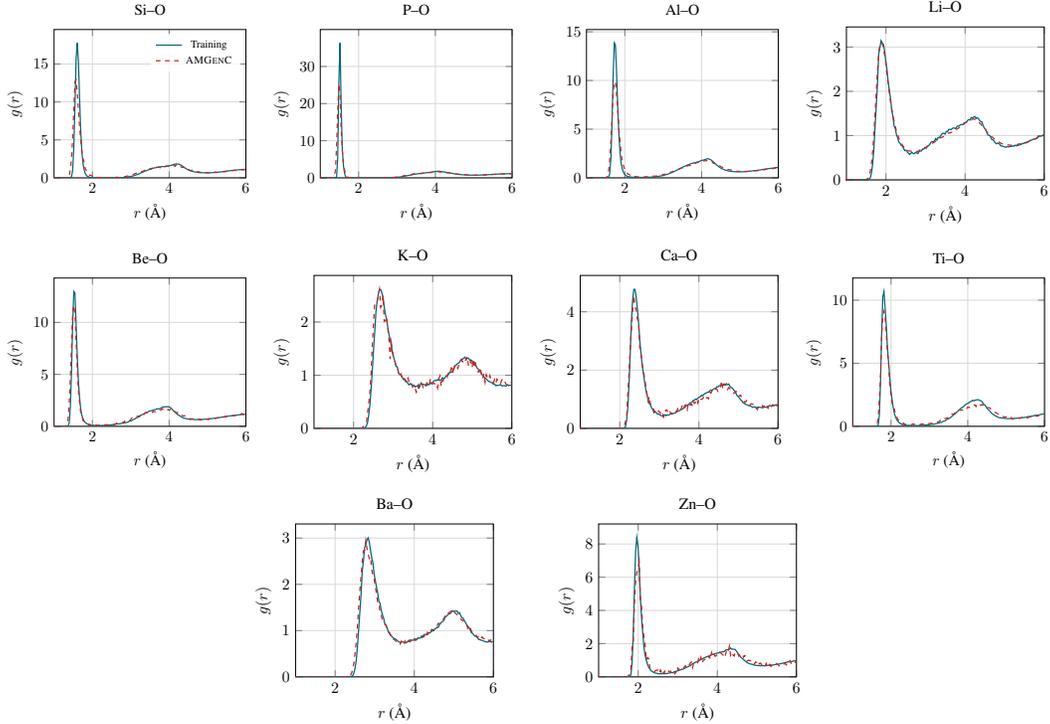

\section{Effect of the Final Discrete Projection}
\label{apx:dp-effect}

The final discrete projection (DP, \Secref{sec:dp}) keeps every atom position fixed and changes only element assignments, so comparing the same generated samples immediately before and after DP isolates its effect.

\paragraph{Changed element assignments.}
Table~\ref{tab:dp-changes} reports the number of element assignments changed by DP in samples generated by \archname{}.
The atom fraction is computed over all atoms of a sample, including ghost atoms.
Across the four configurations, DP changes less than one element assignment per sample on average and at most four in any sample.

\begin{table}[ht]
  \caption{Element assignments changed by DP in samples generated by \archname{}.}
  \label{tab:dp-changes}
  \centering
  \begin{tabular}{l ccc}
    \toprule
    Config & Mean changed & Maximum changed & Mean atom fraction changed \\
    \midrule
    a-\ch{SiO2}~$\mid$~$G$ & 0.63 & 2 & 0.098\% \\
    a-\ch{SiO2}~$\mid$~RSD & 0.59 & 2 & 0.092\% \\
    MEG~$\mid$~$E$+$C_\text{Li}$ & 0.81 & 4 & 0.061\% \\
    MEG~$\mid$~$G$ & 0.77 & 3 & 0.058\% \\
    \bottomrule
  \end{tabular}
\end{table}

\paragraph{Structural metrics.}
Each structural metric compares the distribution over generated samples with that over training samples.
Partial radial distribution functions (RDFs) use 200 bins over 0 to 6\,\AA, and bond angle distributions use 180 bins over 0 to 180$^\circ$.
The first-shell cutoff is the first local minimum after the main peak of the partial RDF of training samples, which also defines the neighbors for bond angles.
We report the Jensen-Shannon divergence (JSD, base 2) and the Wright factor~\citep{wright1993comparison} of the partial RDFs, the MAE of the cumulative coordination number $n(r)$ from zero distance to the first-shell cutoff (CN MAE), and the JSD of the bond angle distributions.
For MEG, all metrics are computed for each cation type $X \in \{\mathrm{Si, P, Al, Li, Be, K, Ca, Ti, Ba, Zn}\}$ paired with oxygen, and averaged over $X$.
All comparisons use the same training samples and cutoffs.
The same metrics are used for Table~\ref{tab:dp-ablation}.

Table~\ref{tab:dp-structure} reports the structural metrics before and after DP.
Every metric changes by at most 0.12\% on a-\ch{SiO2}~$\mid$~RSD and 0.25\% on MEG~$\mid$~$E$+$C_\text{Li}$ relative to its value before DP, showing that DP preserves the measured partial RDF, coordination number, and bond angle distributions.

\begin{table}[ht]
  \caption{Structural metrics of samples generated by \archname{} before and after the final discrete projection (DP), compared with training samples. Lower is better for all metrics.}
  \label{tab:dp-structure}
  \centering
  \begin{tabular}{l l cc}
    \toprule
    Config & Metric & Before DP & After DP \\
    \midrule
    \multirow{4}{*}{a-\ch{SiO2}~$\mid$~RSD}
      & Mean JSD of $g_\mathrm{SiO}(r)$, $g_\mathrm{SiSi}(r)$, $g_\mathrm{OO}(r)$ & 0.01192 & 0.01193 \\
      & Mean Wright factor of $g_\mathrm{SiO}(r)$, $g_\mathrm{SiSi}(r)$, $g_\mathrm{OO}(r)$ & 0.22106 & 0.22118 \\
      & CN MAE of $n_\mathrm{SiO}(r)$ & 0.03759 & 0.03762 \\
      & Mean JSD of $\angle\mathrm{OSiO}$, $\angle\mathrm{SiOSi}$ distributions & 0.00838 & 0.00839 \\
    \cmidrule(lr){1-4}
    \multirow{4}{*}{MEG~$\mid$~$E$+$C_\text{Li}$}
      & Mean JSD of $g_{X\mathrm{O}}(r)$ & 0.01415 & 0.01418 \\
      & Mean Wright factor of $g_{X\mathrm{O}}(r)$ & 0.19952 & 0.19989 \\
      & Mean CN MAE of $n_{X\mathrm{O}}(r)$ & 0.04471 & 0.04482 \\
      & Mean JSD of $\angle\mathrm{O}X\mathrm{O}$ distributions & 0.01829 & 0.01832 \\
    \bottomrule
  \end{tabular}
\end{table}

\paragraph{Composition and inverse design metrics.}
On MEG~$\mid$~$E$+$C_\text{Li}$, DP changes the mean atomic percentage of every element by at most 0.01 percentage points, and its standard deviation across samples by at most 0.02 percentage points.
For example, the Li atomic percentage changes only from 14.44\% to 14.45\%, after the property-conditioned generation has already shifted it from the training mean of 1.98\% toward the 15\% target.
The inverse design MAPE changes from 20.2\% to 20.1\% for $E$ and from 9.9\% to 9.8\% for $C_\text{Li}$ (Table~\ref{tab:ablation}).
The property-conditioned generation thus determines the composition, while DP has a negligible effect on it.

\section{Baseline Adaptation}
\label{apx:baseline-adaptation}

All baselines are retrained on the same datasets and conditioned on the same target properties as \archname{}, with the training and sampling settings in \Secref{sec:experimental-settings}.
The adaptation of each baseline is as follows.

\paragraph{CDVAE.}
Its original workflow first chooses the composition, lattice, and number of atoms, then generates atom positions and elements.
We retain this workflow, retrain it on the amorphous datasets, and replace its optimization of a single property toward an extreme value with matching all target property values.

\paragraph{CrystalFlow.}
Its original workflow generates the lattice, atom positions, and elements jointly while conditioning on numerical properties.
We retain this workflow and retrain it on the amorphous datasets with all target properties as conditioning.

\paragraph{MatterGen.}
Its original workflow first learns material generation from unlabeled structures, then adapts to labeled property targets.
We follow this workflow with unlabeled and labeled amorphous samples, and retain its joint generation of the lattice, atom positions, and elements.

\paragraph{Graphite.}
Its original workflow generates amorphous structures using the error of a XANES spectrum prediction to guide sampling.
We retrain it for multiple elements and change the prediction target from per-atom XANES spectra to the target properties.

\paragraph{AMDEN.}
Its original task and workflow already match our setting, so we retrain it for each dataset and property set.
It generates atom positions and elements jointly inside a fixed cell, using removable empty sites, similar to ghost atoms in this work, to vary the number of atoms and the density.

\section{Experimental Environment}
\label{apx:environment}

\paragraph{Hardware.}
All experiments are conducted on an internal GPU cluster managed by the SLURM workload scheduler.\footnote{\url{https://slurm.schedmd.com/overview.html}}
Each compute node is equipped with two AMD EPYC 7302 16-core processors (32 cores, 64 threads total, base 1.5\,GHz, boost 3.0\,GHz), 256\,GB of system memory, and three NVIDIA A40 GPUs (48\,GB memory each).
Each experiment uses a single A40 GPU, 4 CPU cores, and 64\,GB of system memory.

\paragraph{Software.}
The implementation uses Python~3.12 and PyTorch~2.9.1~\citep{paszke2019pytorch} with CUDA~12.4.
Key dependencies include NumPy~2.4.1~\citep{harris2020array}, SciPy~1.17.0~\citep{virtanen2020scipy}, ASE~3.27.0~\citep{larsen2017atomic}, and pymatgen~2025.10.7~\citep{ong2013python}.
All experiments run inside Singularity~\citep{kurtzer2017singularity} containers built from the \texttt{nvidia/cuda:12.4.1-devel-ubuntu22.04} base image to ensure reproducibility across nodes.
The implementation code and sample datasets are available at \url{https://github.com/Logan-Lin/AMGenC-code}.

\section{Discussion}
\label{apx:discussion}

\subsection{Broader Impacts}
\label{apx:broader-impacts}
\paragraph{Accelerating materials discovery.}
By guaranteeing charge balanced outputs, \archname{} makes generative inverse design of amorphous materials more practical, reducing wasted computation from discarding charge unbalanced samples.
As shown in Table~\ref{tab:efficiency}, post-hoc filtering can require orders of magnitude more generation time to obtain charge balanced samples, whereas \archname{} eliminates this overhead entirely.
This can accelerate the discovery of amorphous materials for applications such as energy storage, thermal management, and advanced materials, where efficient exploration of the vast design space is important.

\paragraph{Potential for broader applications.}
Beyond amorphous materials, the proposed framework for enforcing non-differentiable hard constraints on discrete outputs, through the combination of data-centered conditional element noise, soft linearized projection, and final discrete projection, may generalize to other generative modeling settings where hard constraints on discrete data must be satisfied, such as point cloud generation with categorical label constraints or graph generation with degree-sum invariants.

\subsection{Limitations}
\label{apx:limitation}
\paragraph{Gap between generation and real-world synthesis.}
Generated samples represent computationally predicted atomic configurations, but current experimental techniques cannot directly fabricate amorphous materials atom by atom.
Nevertheless, generated samples are highly informative as candidates that narrow the search space for composition and processing choices, much as traditional computational methods like reverse Monte Carlo~\citep{mcgreevy1988reverse} have long been used to guide materials design.
Additionally, these candidates still require validation through simulation (e.g., molecular dynamics relaxation) or experimental synthesis before real-world deployment.

\paragraph{Scope of the formal charge assignment.}
The guarantee of \archname{} is defined relative to the supplied formal charge assignment.
Formal charge balance is a necessary global condition on the composition of the evaluated oxide glasses (Appendix~\ref{apx:stoichiometry}), but it does not imply electronic neutrality or local chemical validity, which require separate assessments such as bond valence and stability analysis.
For the evaluated datasets, the formal charge assignment is unambiguous, since the samples are constructed from neutral oxide units with fixed oxidation states.
Materials with ambiguous or composition-dependent oxidation states require extensions.
Fixed rational charges can be multiplied by a common denominator and passed to the same discrete projection.
Known mixed-valence states, such as \ch{Fe^{2+}} and \ch{Fe^{3+}}, can be represented as separate element classes when the allowed states are specified in advance.
Composition-dependent charges would require a differentiable charge predictor for the soft linearized projection and an exactly testable discrete charge representation for the final discrete projection.
Evaluating these extensions remains future work.

\paragraph{Structural accuracy.}
This work focuses on enforcing charge balance on element compositions and is not aimed at improving the structural accuracy of generated amorphous materials.
As shown in Appendix~\ref{apx:structural-features}, minor structural discrepancies remain in the partial radial distribution functions of generated samples, where peaks can be lower and broader than those of the training data.
This is consistent with known limitations of diffusion and flow matching models in fully reproducing the well-relaxed local structures of amorphous materials~\citep{finkler2025inverse}.
Improving structural accuracy is an orthogonal direction that could complement the charge balance guarantees provided by \archname{}.

\paragraph{Generalizability.}
While the proposed framework for enforcing hard constraints on discrete outputs may apply to broader generative modeling problems, we evaluate it only in the context of amorphous materials generation across two datasets.
Assessing how well the approach transfers to other domains with different constraint structures remains future work.

\subsection{LLM Usage}
\label{apx:llm-usage}
LLM usage in this work is limited to proofreading the drafted paper and assisting with part of the implementation code, after the methodology and algorithms were designed and specified.
LLMs were not used as a component of the proposed method.

\newpage

\end{document}